\documentclass{article}
\usepackage[final]{neurips_2021}

\usepackage[utf8]{inputenc} 
\usepackage[T1]{fontenc}    
\usepackage{hyperref}       
\usepackage{url}            
\usepackage{booktabs}       
\usepackage{amsfonts}       
\usepackage{nicefrac}       
\usepackage{microtype}      
\usepackage{xcolor}         

\usepackage{amssymb, amsmath, amsfonts, mathtools, dsfont}
\usepackage{graphicx}
\usepackage{amsthm}
\usepackage{txfonts}
\usepackage{csquotes}
\usepackage{subcaption}
\usepackage{algorithm, algorithmic}
\usepackage[inline,shortlabels]{enumitem}
\usepackage{bm}
\usepackage{tikz}

\usepackage{shortcuts}
\usepackage{theorems}

\title{Finding Regions of Heterogeneity in Decision-Making via Expected Conditional Covariance}

\author{%
  Justin Lim\thanks{Equal contribution}\\
  MIT CSAIL and IMES \\
  Cambridge, MA \\
  \texttt{justinl@mit.edu} \\
  \And
  Christina X Ji$^*$\\
  MIT CSAIL and IMES \\
  Cambridge, MA \\
  \texttt{cji@mit.edu} \\
  \And
  Michael Oberst$^*$\\
  MIT CSAIL and IMES \\
  Cambridge, MA \\
  \texttt{moberst@mit.edu} \\
  \And
  Saul Blecker\\
  NYU Langone \\
  New York, NY \\
  \texttt{saul.blecker@nyulangone.org} \\
  \And
  Leora Horwitz\\
  NYU Langone \\
  New York, NY \\
  \texttt{leora.horwitz@nyulangone.org} \\
  \And
  David Sontag\\
  MIT CSAIL and IMES \\
  Cambridge, MA \\
  \texttt{dsontag@csail.mit.edu}
}

\begin{document}

\maketitle

\begin{abstract}
Individuals often make different decisions when faced with the same context, due to personal preferences and background. For instance, judges may vary in their leniency towards certain drug-related offenses, and doctors may vary in their preference for how to start treatment for certain types of patients. With these examples in mind, we present an algorithm for identifying types of contexts (e.g., types of cases or patients) with high inter-decision-maker disagreement. We formalize this as a causal inference problem, seeking a region where the assignment of decision-maker has a large causal effect on the decision. Our algorithm finds such a region by maximizing an empirical objective, and we give a generalization bound for its performance. In a semi-synthetic experiment, we show that our algorithm recovers the correct region of heterogeneity accurately compared to baselines. Finally, we apply our algorithm to real-world healthcare datasets, recovering variation that aligns with existing clinical knowledge.
\end{abstract}

\section{Introduction}%
\label{introduction}

Understanding heterogeneity in decision-making is an established problem in medicine \citep{birkmeyer2013understanding, corallo2014systematic, de2006variation}, consumer choice \citep{Ortega2011, Scarpa2005}, and law \citep{Kang2012, Kleinberg2018-ni, Arnold2018-sw}.  In the context of medicine, this is referred to as the study of practice variation \citep{Atsma2020, cabana1999don}, where it is often observed that doctors, facing the same clinical context, will make different decisions.  Likewise, in a legal context, judges often differ in their leniency in their decisions regarding bail \citep{Kleinberg2018-ni}, juvenile incarceration \citep{Aizer2013-ex}, the use of alternatives to incarceration \citep{Di_Tella2013-sp}, and incarceration length \citep{Kling2006-qm}. In some scenarios this variation may be justified: The best medical treatment may not be obvious. In others, it may be grossly unfair, as in the case of racial bias in bail decisions \citep{Arnold2018-sw}.

In this work, we tackle the question of how to find and characterize this variation in the first place.  In particular, we present a learning algorithm for identifying a \enquote{region of heterogeneity}, defined as a subset of all contexts (e.g., patients, cases) for which the identity of the decision-maker substantially affects the decision. In medicine, a better understanding of treatment variation can inform the development and dissemination of clinical guidelines. In the legal domain, characterizing the cases where judges vary most in their leniency may help with investigating potential issues of fairness. 

We formalize characterizing the region of heterogeneity as a causal inference problem: We want to characterize examples where \textit{changing the decision-maker} would have resulted in a different decision. The challenge is two-fold: First, we only observe a single decision-maker per example, so we cannot directly observe how (for instance) multiple judges would have decided the same case. Second, our data on individual decision-makers is often scarce. For instance, in Section~\ref{real-experiments}, we consider a medical dataset with more than 400 doctors, each of whom has fewer than 9 samples on average.

We will refer to decision-makers as \enquote{agents}, and our contributions are as follows: In Section~\ref{sec:causal_perspective}, we propose an objective defined in terms of counterfactual decisions across different agents, and show that this objective can be identified from observational data.  Moreover, this objective does not require the use of agent-specific statistical models, making it amenable to our sparse setting.  In Section~\ref{sec:approach}, we give an iterative algorithm to identify regions of disagreement by maximizing this objective and provide intuition (in the form of a generalization bound) for the factors that drive its performance.  In Section~\ref{sec:semi_synthetic_experiments}, we use a semi-synthetic dataset, derived from crowd-sourced recidivism predictions, to demonstrate that our algorithm recovers the correct region of heterogeneity accurately, even when there are many agents.  Finally, in Section~\ref{sec:diabetes}, we apply our algorithm to a real-world healthcare dataset and confirm that it recovers intuitive regions of variation in first-line diabetes treatment.  We conclude with a discussion of related work and implications. Our code is available at \url{https://github.com/clinicalml/finding-decision-heterogeneity-regions}.

Our algorithm does not determine whether variation is inherently good or bad or how it should be addressed. Rather, more careful study with domain experts would be required to determine if variation can (or should be) reduced and how. In addition, false discovery of variation is possible and could have a negative impact. We expect that validation on independent datasets would be required in real-world applications, using the regions identified by our method as plausible hypotheses to test.

\section{Characterizing Heterogeneity from a Causal Perspective}%
\label{sec:causal_perspective}

\subsection{Notation} 
Let the data be drawn from a distribution $\P(X, A, Y)$, where 
$X$ is a random variable representing context (or features), $A$ is a discrete agent, and $Y \in \{0, 1\}$ is the binary decision.  The spaces of all $X$ and $A$ are denoted as $\cX$ and $\cA$ with realized values as lower case $x$ and $a$, respectively. Indicator variables $\1{\cdot}$ are one if the statement inside the brackets is true and zero otherwise.  For a subset $S \subseteq \cX$, $\1{x \in S}$ is sometimes written as a function $S(x)$, where $S: \cX \rightarrow \{0, 1\}$. A subset $S$ may have several disjoint regions. $\mathbb{E}\left[Y \vert X \in S\right]$ denotes the average $Y$ across samples in $S$. For instance, if $S = \left\{X: X_0 < 10\right\}$, then $\mathbb{E}\left[Y \vert X_0 < 10\right]$ is a scalar average of $Y$ among samples with $X_0 < 10$.

\subsection{Heterogeneity as a Causal Contrast} 

Our conceptual goal is to identify a region $S \subseteq \cX$ where different agents tend to make different decisions even when faced with the same context.  We can formalize this in the language of potential outcomes from the causal inference literature \citep{Pearl2009-sj, Hernan2020-zj}, which for clarity we will refer to as \textit{potential decisions}:  In particular, we denote $Y(a)$ to be the potential decision of agent $a$.  The fundamental challenge of causal inference is that we do not observe all potential decisions $\{Y(a) : a \in \cA\}$ for each sample, but only a single decision $Y$.  With this in mind, we will make the following assumptions, standard in the literature on causal effect estimation.
\begin{thmasmp}[Causal Identification Assumptions]\label{thmasmp:causal_identification}
\begin{enumerate*}[label=(\roman*)]
    \item Consistency: $Y = y, A = a \implies Y(a) = y$, and 
    \item No Unmeasured Confounding (NUC): For all $a \in \cA$, $Y(a) \indep A \mid X$.
\end{enumerate*}
\end{thmasmp}
Consistency links the potential $Y(a)$ to the observed $Y$, and NUC says that there are no unobserved factors that influence both the assignment of agents and the decision itself.  For instance, the quasi-random assignment of cases to judges conditioned on features $X$ satisfies NUC \citep{Kleinberg2018-ni}. NUC may be violated if key aspects of the case (e.g., misdemeanor vs.\ felony) are omitted as features.  For instance, misdemeanor and felony cases may be seen by different judges and have different decision processes, but this variation is not due to agent preferences. Given these assumptions, we propose a causal measure of agent-specific bias, defined as a contrast between potential decisions.
\begin{thmdef}[Conditional Relative Agent Bias]\label{thmdef:conditional_relative_agent_bias}
  For an agent $a \in \cA$ and a subset $S \subseteq \cX$, the conditional relative agent bias is defined as 
  \begin{equation}\label{eq:conditional_relative_agent_bias}
    \E[Y(a) - Y(\pi(x)) \mid A = a, X \in S] 
  \end{equation}
  where $Y(a)$ is the potential decision of agent $a$, and $Y(\pi(x)) \coloneqq \sum_{a'} \E[Y(a') \mid x] \pi(a' \mid x)$ denotes the expected potential decision under the agent assignment distribution $\pi(a' \mid x) \coloneqq \P(A = a' \mid X = x)$.
\end{thmdef} 
Note that under Assumption~\ref{thmasmp:causal_identification}, $Y(\pi(x)) = \E[Y \mid x]$,\footnote{See Proposition~\ref{prop:ypi_conditional_exp} for a short proof, and Proposition~\ref{prop:rewrite_conditional_relative_agent_bias} for the derivation of Equation~\eqref{eq:conditional_relative_agent_bias_expanded}.}
but here we emphasize its causal interpretation as the expected decision of a random agent. Equation~\eqref{eq:conditional_relative_agent_bias} represents the relative difference between the decision of an agent (on their particular distribution of cases in the region) and the potential decision of a random agent. In particular, Equation~\eqref{eq:conditional_relative_agent_bias} can be written as follows under Assumption~\ref{thmasmp:causal_identification}
\begin{equation}\label{eq:conditional_relative_agent_bias_expanded}
\E[Y(a) - Y(\pi) \mid A = a, X \in S] = \int_{x \in S} \E[Y(a) - Y(\pi) \mid X = x] p(x \mid A = a, X \in S) dx,
\end{equation}
where we shorten $Y(\pi(x))$ to $Y(\pi)$. This is the average difference (over $p(x \mid a)$, restricted to those $x$ in the set $S$) of the conditional expected difference between $Y(a)$ and $Y(\pi)$. For example, suppose that the agent $a$ is a judge who is particularly lenient on bail decisions for felony arrests (the region $S$), and $Y = 1$ denotes granting bail.  Then imagine the following counterfactual: Take the felony cases that are assigned to this judge and reassign each individual case, described by $x$, to a random judge $a'$, proportionally to $p(a' \mid x)$.  We may then observe, on average, that the bail rate would decrease, because most judges are less lenient than judge $a$, corresponding to a positive value of Equation~\eqref{eq:conditional_relative_agent_bias}.

Equation~\eqref{eq:conditional_relative_agent_bias} has the additional advantage of being easy to estimate:  Under Assumption~\ref{thmasmp:causal_identification}, it can be rewritten\footnote{See Proposition~\ref{prop:crab_as_residuals} in Appendix~\ref{sec:proof_ypi}.} as $\E[Y - \E[Y \mid X] \mid A = a, X \in S]$, the expected residual in predicting (using the conditional expectation $\E[Y \mid X]$) the decision of an agent $a$ across the context $x$ typically seen by that agent.

\subsection{Formalizing a Causal Objective}%
\label{sec:objective_assumptions}

Our primary goal is to discover a region $S$ where substantial heterogeneity exists across agents.  To do so, we define an aggregate objective across a group $G$ of agents, where $G(a) \in \{0, 1\}$ is an indicator function for membership.
\begin{align}
  Q(S, G) \coloneqq \sum_{a: G(a) = 1} \P(A = a \mid X \in S) \E[Y(a) - Y(\pi) \mid A = a, X \in S],
\end{align}
We now show that this quantity can be identified and estimated from observational data without requiring agent-specific statistical models, before discussing the interpretation of this objective.
\begin{thmthm}[Causal Identification]\label{thmthm:causal_objective}
Under Assumption~\ref{thmasmp:causal_identification}, $Q(S, G)$ can be identified as 
\begin{align}
  Q(S, G) = \E_S[\cov(Y, G \mid X)] = \E_S[(Y - \E[Y \mid X]) G],\label{eq:exp_cond_cov}
\end{align}
where $\E_{S}[\cdot] \coloneqq \E[\cdot \mid X \in S]$ and $\cov(Y, G \mid X)$ is the conditional covariance.
\end{thmthm}
Theorem~\ref{thmthm:causal_objective} and other theoretical results are proven in Appendix~\ref{sec:proofs}. The result follows from proving that the agent-specific bias (Definition~\ref{thmdef:conditional_relative_agent_bias}) is identifiable using the expected conditional covariance between $Y$ and the binary indicator $\1{A = a}$. With this in mind, we optimize the following objective, where  the set $S$ is constrained to be at least a certain size $\beta$ and $\cS$ is a hypothesis class of functions $S$.
\begin{align}
    \max_{S \in \cS, G}  Q(S, G)\ \text{s.t.,} \ \P(S) \geq \beta, \label{eq:objective}
\end{align}

\textbf{Interpretation}: Intuitively, this objective measures the disagreement between the agents in the group $G(a) = 1$ and the overall average $\E[Y\mid X]$ on the region $S$.  Hence, the choice of group is important for interpreting the objective: If $G(a) = 1$ for all agents, the objective will be zero for any set $S$, as can be seen from Equation~\eqref{eq:exp_cond_cov}, applying the definition of the conditional expectation.  

Accordingly, we seek a region $S$ for which the partially maximized objective $L(S) \coloneqq \max_G Q(S, G)$ is large: This partial maximization is obtained by taking $G(a) = 1$ whenever the conditional relative agent bias of agent $a$ (on the set $S$) is non-negative. Thus, Equation~\eqref{eq:objective} can be re-written as
\begin{align}
  \max_{G} Q(S, G) = \sum_{a \in \cA} \P(A = a \mid X \in S) \abs{\E[Y(a) - Y(\pi) \mid A = a, X \in S]}_{+}, \label{eq:rewriting_abs_values}
\end{align}
where $\abs{x}_{+} \coloneqq \max(x, 0)$, and this objective becomes an average over agents who have a positive bias. This population objective is also equivalent (up to a constant factor) to the (weighted) sum of the magnitude of each agent's conditional relative agent bias. See Proposition~\ref{prop:equivalence_to_abs_values} in Appendix~\ref{sec:proof_ypi}.

\textbf{Lack of Overlap}: We have \textit{not} made the overlap or positivity assumption that $\P(A = a \mid x) > 0$ for all $x, a$.  While this assumption is required to identify conditional causal effects $\E[Y(a) - Y(a') \mid X]$ \citep{nie2017quasi, wager2018estimation, shalit2017estimating}, it is not required for identifying our causal contrast. Our problem only requires each context has a positive probability of being seen by more than one decision maker. For instance, suppose that $S$ contains both misdemeanors and felonies and there are four judges $a_0, \ldots, a_3$. If judges $a_0$ and $a_1$ make bail decisions exclusively for felonies while judges $a_2$ and $a_3$ make bail decisions exclusively for misdemeanors, our measure captures disagreement between $a_0$ and $a_1$ and between $a_2$ and $a_3$ even though comparisons between $a_0$ and $a_2$ or other pairs are impossible to make. Thus, we have chosen to compare $Y(a)$ to the decisions of \textit{viable alternative agents}, weighted by their probability $p(a' \mid x)$ of being selected for such a case.

\section{Identifying Regions with Heterogeneity} 
\label{sec:approach}

In Section~\ref{sec:algorithm}, we introduce an iterative optimization algorithm for a finite sample version of Objective~\eqref{eq:objective} that alternatingly optimizes $S$ and $G$. In Sections~\ref{sub:tuning_the_region_size} and~\ref{sub:hypothesis_test_validation}, we discuss practical heuristics for choosing the region size parameter $\beta$ on training data and validating if the resulting region generalizes to held-out data.  Finally, we build intuition for the factors that influence performance of this algorithm via a generalization bound in Section~\ref{sec:algorithm_analysis} under simplifying assumptions.

\subsection{Iterative Optimization Algorithm}%
\label{sec:algorithm}

We let $\hat{Q}(S, G)$ be the empirical analog of $Q(S, G)$ (Equation~\ref{eq:exp_cond_cov}), which we can write as follows, 
\begin{align}
\hat{Q}(S, G) &\coloneqq \frac{1}{\sum_{a,j} \1{x_{aj} \in S}} \sum_{a, j} (y_{aj}-f(x_{aj}))\cdot G(a) \cdot \1{x_{aj} \in S}. \label{eq:empirical_expected_conditional_covariance}
\end{align}
where $f(x)$ denotes a model of the conditional expectation $f(x) \approx \E[Y \mid X = x]$.  For simplicity of notation, we assume that there are $R$ samples (indexed by $j$) for each of a finite set of $N$ agents (indexed by $a$), giving $N \cdot R$ samples in total. 

Our algorithm (Algorithm~\ref{alg:main}) takes as input the data $\{(x_{aj}, y_{aj})\}$ and a minimum region size $\beta$, and outputs a model $h(x)$ and a threshold value $b$ that describe a region of heterogeneity $S=\{x\in \cX; h(x)\geq b\}$. Starting with $S = \cX$ (the entire space), the algorithm identifies the grouping that maximizes $\hat{Q}(S, G)$, then uses that grouping to identify the region maximizing the same quantity, repeating this process until convergence. The algorithm uses a classifier $f(x)$ to estimate $\E[Y\mid X=x]$ and a regression model $h(x)$ to estimate the conditional covariance of the decision $Y$ and the grouping $G$ at $X=x$. Note that we can use any supervised learning algorithms for $f$ and $h$, allowing us to learn interpretable regions as part of the algorithm if $h(x)$ is interpretable (e.g., decision trees). If sufficient data is available, samples can be split into three parts for estimating $f(x)$ in line 2, computing $G(a)$ in lines 5-8, and training $h(x)$ and estimating the $(1-\beta)$-th quantile in line 10. We do not perform this sample splitting because our sample sizes are already small. Under-fitting $f(x)$ by further restricting the sample size could lead to false discovery if $f(x)$ does not capture the variation explained by $X$.

\textbf{Optimizing over $G$ given $S$.} Given a region $S$, our first result identifies the grouping $G:\cA\rightarrow \{0, 1\}$ that maximizes $\hat{Q}(S, G)$ and shows that it can be expressed in terms of $\hat{Q}(S, \1{A=a})$.

\begin{thmprop}\label{thmprop:optimal_partition}
Given $S\subseteq \cX$, $\hat{Q}(S, G)$ is maximized over the space of functions $G:\cA\rightarrow \{0, 1\}$ at $G_S$, where $G_S(a) = \1{\hat{Q}(S, \1{A=a})\ge 0}$.
\end{thmprop}
Intuitively, this proposition states that to maximize the empirical expected conditional covariance of the decision and grouping on a region, we must group agents by whether their residuals $y_{aj}-f(x_{aj})$ are (on average) positive or negative on $S$. 

\textbf{Optimizing over $S$ given $G$.} 
To optimize $\hat{Q}(S, G)$ for a fixed grouping $G$ over the hypothesis class $\cS$, we train a model $h(x)$ to predict $(y_{aj}-f(x_{aj}))G(a)$ given $x_{aj}$, where $h\in \cH$. Using $h$ as an estimate in Eq.~\ref{eq:empirical_expected_conditional_covariance}, we find a set $S$ to maximize the quantity $\frac{1}{\sum_{x}\1{x \in S}} \sum_{x \in S} h(x)$, the empirical conditional expectation of $h(x)$ over $S$. This quantity is maximized (subject to our $\beta$ constraint) by including the largest $\beta$-fraction of the $h(x_{aj})$ in $S$. Hence, we pick $b$ as the $(1-\beta)$-th quantile of $h(x_{aj})$ and choose our region as $\hat{S}_G=\{x\in \cX; h(x)\geq b\}$.

\begin{algorithm}[t]
    \caption{Identifying regions with variation}\label{alg:main}
\begin{algorithmic}[1]
\STATE {\bfseries Input:} Data ${{\{\{x_{aj}, y_{aj}\}}_{j=1}^R\}}_{a=1}^N$, minimum region size $\beta$.
    \STATE Fit a model $f(x)$ to $\E(Y\mid X=x)$. 
    \STATE Initialize $S=\cX$.
    \REPEAT
        \FOR{$a=1, \dots, N$}
            \STATE Compute $\hat{Q}(S, \1{A=a})$, where $\hat{Q}(S, \1{A=a})\coloneqq \frac{1}{\sum_{j}\1{x_{aj}\in S}}\sum_{j} (y_{aj}-f(x_{aj})) \1{x_{aj} \in S}$,
            \STATE Set $G(a)=1$ if $\hat{Q}(S, \1{A=a})\ge 0$ and $0$ otherwise.
        \ENDFOR
        \STATE Compute $b_{aj}=(y_{aj}-f(x_{aj}))G(a)$, $a=1, \dots, N$, $j=1, \dots, R$.
        \STATE Fit a model $h(x)$ to predict $b_{aj}$ from $x_{aj}$, and let $b$ be the $(1-\beta)$-th quantile of $h(x_{aj})$.
        \STATE $S'\leftarrow S$.
        \STATE $S\leftarrow \{x_{aj};\ h(x_{aj})\geq b\}$.
    \UNTIL{$S=S'$ or iteration limit reached.}
    \STATE \textbf{Output:} Model $h$ and threshold $b$, defining a region $S=\{x\in \cX; h(x)\geq b\}$.
\end{algorithmic}
\end{algorithm}

\subsection{Tuning the Region Size Parameter}%
\label{sub:tuning_the_region_size}

For real datasets, we need to choose $\beta$ without knowledge of the true value. Given that our objective can be calculated on held-out data using the functions $S, G$, a seemingly obvious approach would be to compute $Q(S, G)$ on a validation set and select the $\beta$ that leads to the highest $Q(S, G)$. However, for a fixed data distribution, smaller values of $\beta$ tend to produce higher values of $Q(S, G)$, and there is a trade-off between finding a smaller region of higher variation and a larger region that may include areas of lower (but still meaningful) variation. This motivated our original constraint $\P(S) \geq \beta$.

To select $\beta$, we propose a heuristic inspired by permutation-based hypothesis testing \citep{Wasserman2004-qm}. We compare the training objective to a reference distribution of values (for the same $\beta$) that might be seen \textit{even if all agents followed the same policy}. For each candidate $\beta$, we \begin{enumerate*}[label=(\roman*)]
  \item run our algorithm and compute the objective on training data $q_{\text{obs}} \coloneqq \hat{Q}(\hat{S}, \hat{G})$. 
  \item For $T$ iterations, we randomly shuffle the agents and re-run the algorithm to get a new objective value.  This gives us a distribution $\hat{\P}_{\text{null}}$ over $Q(S, G)$ from a distribution where $\P(X, Y)$ and $\P(A)$ are unchanged but $Y, X \indep A$.
  \item Finally, we compute a p-value $p_{\beta} = \hat{\P}_{\text{null}}(Q > q_{\text{obs}})$ and choose the $\beta$ with the smallest p-value.
\end{enumerate*}
In Section~\ref{sub:semisynth_tuning_beta}, we find that this heuristic empirically recovers the true $\beta$ value in semi-synthetic settings.

\subsection{Validation of the Region}%
\label{sub:hypothesis_test_validation}

We may wish to validate the learned region $\hat{S}$ independently of the grouping $\hat{G}$.  In particular, finding $\hat{G}$ is not our main goal, and we observe in our semi-synthetic experiments that our algorithm can find the true region $S$ even when the grouping $\hat{G}$ is fairly poor (due to few samples per agent), as shown in Appendix~\ref{sec:baseline_app}. We can optimize over $G$ in $Q(S, G)$ to obtain an objective that depends only on $S$ and can be used to compare regions. By Proposition~\ref{thmprop:optimal_partition}, we obtain an empirical analog of Equation~\eqref{eq:rewriting_abs_values} as
\begin{equation}
    \hat{L}(S) \coloneqq \max_{G} \hat{Q}(S, G) = \frac{1}{\sum_{a,j} \1{x_{aj} \in S}}\sum_{a} \left|\sum_{j} (y_{aj}-f(x_{aj}))\1{x_{aj} \in S}\right|_{+}, \label{eq:partial_optimization}
\end{equation}
where $|x|_{+}$ is equal to the positive part $[x]_{+} = \max(x, 0)$ as before. We then use this objective $\hat{L}(S)$ to answer the following question:  Does our chosen region $\hat{S}$ yield a significantly higher objective value on test data than a randomly selected region of the same size?  An example of this analysis is given in Table~\ref{tab:IterativeAlgDecisionTree_benchmark_diabetes} for the real-data experiment in Section~\ref{sec:diabetes}.

\subsection{Generalization Error} 
\label{sec:algorithm_analysis}

We give a generalization bound for Algorithm~\ref{alg:main} to build intuition for the factors that influence performance.  To derive this bound, we consider a simplified setting, where there exists a set $S', G'$ such that the following set of assumptions hold.
\begin{thmasmp}[Group-based variation]\label{assumption:zero_outside_s}\label{assumption:homogeneity} 
For all $x \in S'$, $\E[Y\mid X=x, A = a] = \E[Y\mid X=x, G'(a)]$ and for all $x \not\in S'$, $\E[Y\mid X=x, A = a] = \E[Y\mid X=x]$
\end{thmasmp}
\begin{thmasmp}[Non-zero relative biases]\label{assumption:separable}
There exists a constant $\alpha>0$ such that for all $x\in S'$, 
\begin{align*}
  \E[Y \mid X = x, G'(A) = 1] - \E[Y\mid X = x] &> \alpha,  & & \text{and} &  \E[Y \mid X = x, G'(A) = 0] - \E[Y\mid X = x] &< -\alpha, 
\end{align*}
\end{thmasmp}
\begin{thmasmp}[All agents see samples in $S'$]\label{assumption:coverage}
There exists a constant $\omega > 0$, such that for every $a \in \cA$, $\P(X \in S' \mid A = a) > \omega \P(X \in S')$.
\end{thmasmp}
Note that under these assumptions, $S', G'$ maximize the objective $Q(S, G)$ (see Appendix~\ref{app:s_maximizes_objective}), so we will refer to them as $S^*, G^*$ for the remainder of this section.  Assumption~\ref{assumption:homogeneity} says that there are two groups of agents, who follow two distinct decision policies within a region $S^*$ but follow an identical decision policy outside of $S^*$.  Assumption~\ref{assumption:separable} says that one group has a positive bias across all of $S^*$, relative to the average over both groups, and the other group has a negative bias. To simplify the analysis, we also make Assumption~\ref{assumption:coverage} that every agent has some non-zero chance of observing some contexts $X$ in the region, but note that we do not require that $p(x \mid a) > 0$ for all $x \in S'$. Under these assumptions, we demonstrate that the first iteration of Algorithm~\ref{alg:main} will find, with high probability, a region $\hat{S}$ whose value $Q(\hat{S}, G^*)$ (for the same grouping $G^*$ defined above) is close to that of the optimal $S^*$. Note that we do not claim that the iterative algorithm finds the globally optimal solution. For simplicity, we assume that $f(x)$ perfectly recovers $\E[Y \mid X]$. This can be relaxed at the cost of additional terms in the bound that go to zero as the overall sample size increases. Under Assumptions~\ref{assumption:homogeneity},~\ref{assumption:separable}, and~\ref{assumption:coverage}, assume that $\P(S^*) = \beta$. For the informal version presented here, we assume that $\P(S^*) = \P(\hat{S}) = \beta$, where $\hat{S}$ is returned by our algorithm, and that exactly a $\beta$-fraction of our samples fall into $S^*$ and $\hat{S}$ (in the Appendix, we give a version without these simplifications).
\begin{thmthm}\label{thm:correctness}
Under the assumptions above, if $S^* \in \cS$ and $R > \frac{2 \ln 2}{\alpha^2 \beta^2 \omega^2}$, the first iteration of Algorithm~\ref{alg:main} returns $\hat{S}$ such that, with probability at least $1-\delta$, $Q(S^*, G^*)-Q(\hat{S}, G^*)\le \epsilon$, where
\begin{align*}
   \epsilon &= \sqrt{\frac{2\ln(3/\delta)}{\beta N\cdot R}}+\frac{2}{\beta} \left(\eta+\sqrt{\frac{3\eta(1-\eta)}{\delta \cdot N}}\right) + \frac{1}{\beta}\left( 2 \cR(\cS, N\cdot R) + 4 \sqrt{\frac{2 \ln (12 / \delta)}{N \cdot R}} \right),
\end{align*}
where $\cR(\cS, N \cdot R)$ is the Rademacher complexity of $\cS$, and $\eta \coloneqq \exp\left(-R\alpha^2\beta^2\omega^2 / 2\right)$.

\end{thmthm}

The term $\eta$ plays an important role:  It bounds the expected misclassification error $\P(\hat{G}(a) \neq G^*(a))$. For sufficiently large $R$, we have that $\eta < 1/2$ with high probability, i.e., we have a better-than-random chance of identifying the group for an individual agent.  Moreover, $\eta$ decreases as we increase the number of samples $R$ for each agent, the separation $\alpha$ between the two groups on $S^*$, the region size $\beta$, and the constant $\omega$.  For sufficiently small $\eta$, our algorithm discovers a region whose value (in terms of $Q(S, G^*)$) is close to that of the optimal region. 
The generalization bound improves as the number of agents $N$ increases, the number of samples $R$ for each agent increases, or the complexity of the hypothesis class decreases. The latter is measured here by the Rademacher complexity $\cR(\cS, N \cdot R)$ of our hypothesis class $\cS$, which can be bounded by standard arguments. In conclusion, under some additional assumptions, Algorithm~\ref{alg:main} identifies an approximately optimal solution with high probability after one iteration. We show via semi-synthetic experiments in Appendix~\ref{conv-semisynth-app} that convergence is generally fast in practice.

\section{Semi-Synthetic Experiment: Recidivism Prediction}%
\label{sec:semi_synthetic_experiments}

For conceptual motivation in the introduction, we discussed the legal system: As a potential application of our method, one could determine types of cases for which the idiosyncratic preferences of judges have a significant impact on their decisions.  Lacking data on judge decisions with sufficient context, we turn to the more controlled setting of human predictions of recidivism.

\textbf{Dataset:} We use publicly available data from \citet{Lin2020-ev},\footnote{Available at \url{https://github.com/stanford-policylab/recidivism-predictions}} who ask participants on Amazon's Mechanical Turk platform to make recidivism predictions based on information present in the \enquote{Correctional Offender Management Profiling for Alternative Sanctions} (COMPAS) dataset for Broward County, FL \citep{Dressel2018-br}. Participants (or \enquote{agents}) 
are shown 5 risk factors: age, gender, number of prior convictions, number of juvenile felony charges, and number of juvenile misdemeanor charges.  The charge in question is also given, as well as whether the charge is a misdemeanor or felony.  The dataset contains 4550 cases evaluated by 87 participants.

\textbf{Semi-Synthetic Policy Generation:} To benchmark our method, we generate semi-synthetic data where we have access to a \enquote{ground truth} region of heterogeneity.  We retain the features presented to the original participants and construct two policies, which we refer to as the \enquote{base} and \enquote{alternative} policies: For the base policy, we learn a logistic regression model on the binary decisions across the whole dataset. For the alternative policy, an extra positive term is added to the logistic regression for samples within the region. We construct two scenarios with different regions of variation:  (1) all drug possession charges, and (2) all misdemeanor charges where the individual is 35 years old or younger.  These make up 22\% and 21\% of the data, respectively. Then, we generate synthetic agents (randomly assigned to cases) and assign half of the agents to the base policy and half to the alternative.  Synthetic decisions are then sampled from the logistic regressions.  For each scenario, the two groups of agents follow the same stochastic policy outside of the region, and one group systematically prefers $Y = 1$ within the region. More details can be found in Appendix~\ref{sec:semisynth-setup-app}.

\subsection{Performance versus Baselines}%
\label{sub:performance_versus_baselines}
\begin{figure}[t]
\centering
    \includegraphics[width=0.9\textwidth]{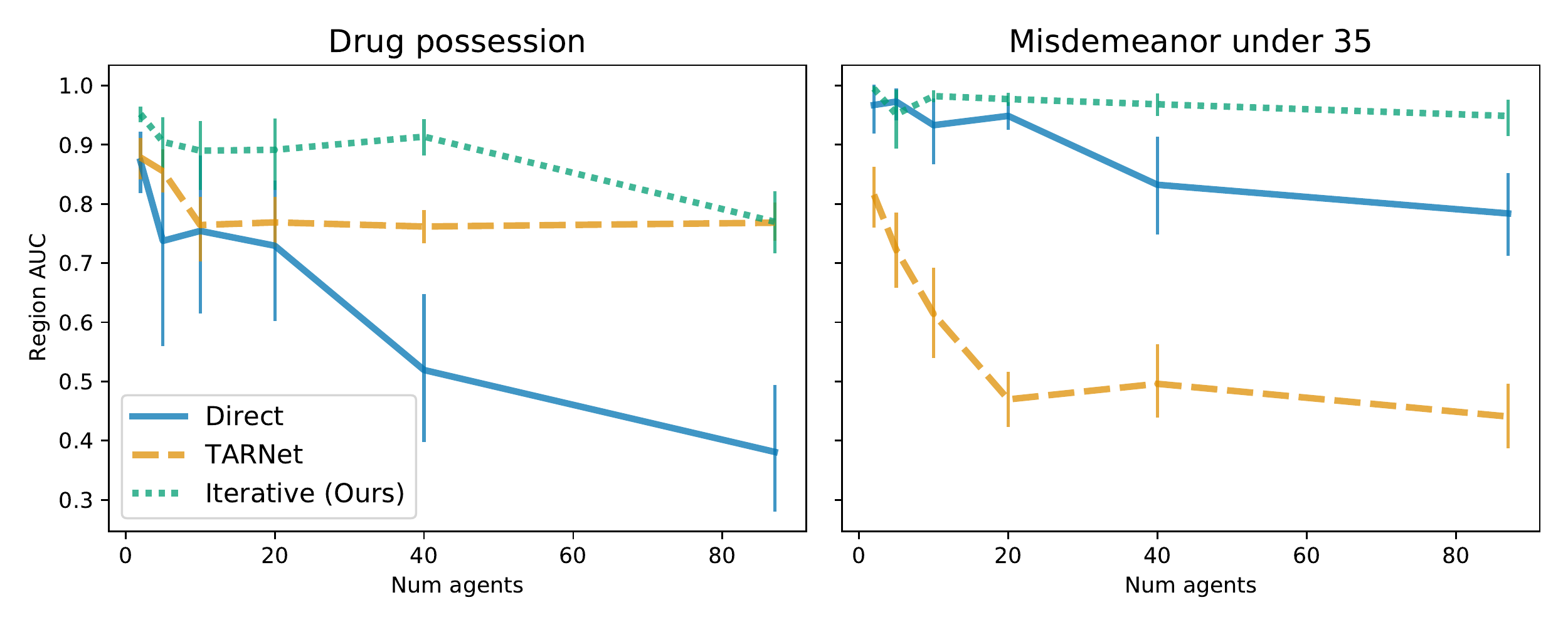}
    \caption{Comparison of our method and best baselines at identifying region of heterogeneity, as measured by the held-out test AUC for classifying samples into the true region of heterogeneity. Total number of samples is fixed.  Baselines are described in Section~\ref{sub:performance_versus_baselines}. Uncertainty bands give 95\% intervals for the mean derived via bootstrapping over 10 random seeds using seaborn \citep{Waskom2021seaborn}. Left: Region is modelled using a ridge regression in the drug possession semi-synthetic set-up. Right: Region is modelled using a random forest for the misdemeanor under age 35 set-up.}%
    \label{fig:semisynth_region_auc}
\end{figure}

\textbf{Baselines:} We compare how well our approach identifies the true region of heterogeneity with several baselines. To our knowledge, the problem of finding regions of heterogeneity with a large number of agents has not been studied before. Many causal inference methods for treatment effect estimation are designed for a single, binary treatment.  However, naively estimating the treatment effect between each pair of providers would scale $O\left(\vert \mathcal{A} \vert^2 \right)$. Therefore, we develop new baselines. Some (including the causal forest and U-learner adaptations described in Appendix \ref{sec:baseline_app}) are based on causal inference methods augmented to identify a region of heterogeneity and grouping of agents where possible.

\textit{Direct models:} This baseline measures how much adding the agent to the feature set improves prediction of decisions.  We fit logistic regressions with and without the agent feature to estimate $\E[Y\mid A, X]$ and $\E[Y\mid X]$.  For each sample $(x, y, a)$, we compute $\vert y - \E[Y\mid X = x]\vert - \vert y - \E[Y\mid A = a, X = x] \vert$ to quantify how much the model with agents outperforms the model without agents. Then, we fit a \enquote{region model} to predict this quantity from $X$.  This region model is either a ridge regression, decision tree, or random forest model. Finally, we compute the top $\beta$ quantile of predictions from the region model in the training and validation sets and use this cut-off to select points in the test set.

\textit{TARNet:} A treatment-agnostic representation network \citep{shalit2017estimating} models the outcomes of all treatments for each sample by learning a shared representation of the sample features and then having a separate prediction head for each treatment. We implement the shared representation model using a 2-layer neural network with ReLU and dropout. Each prediction head is a linear layer with a sigmoid function. TARNet predicts $\E[Y\mid X,A]$ for every agent for each sample. $\var_A[\E[Y\mid X,A]]$ measures the variation across all agents if they had seen context $X$ and is analogous to our objective without grouping. We predict this quantity with the region models as in the direct model baseline.

\textbf{Results:} We evaluate how well the algorithms identify the samples within the region of heterogeneity when we vary the number of agents among 2, 5, 10, 20, 40, and 87, where 87 is the number of agents in the original dataset. Figure~\ref{fig:semisynth_region_auc} shows the best overall region models for each set-up, with the other models deferred to Appendix \ref{sec:baseline_app}. The metric in Figure~\ref{fig:semisynth_region_auc} is the region AUC, defined as how well the model classifies whether samples belong in the region of heterogeneity when compared to the true region. Algorithm~\ref{alg:main} consistently performs well for both semi-synthetic set-ups,  especially when the number of agents increases to a realistic level (and the number of samples per agent decreases). The direct baseline deteriorates very rapidly as the number of agents increases in the drug possession set-up, while the TARNet baseline deteriorates rapidly in the misdemeanor under age 35 set-up. Refer to Appendix \ref{sec:baseline_app} for additional baseline details, region models, and evaluation metrics. We also show that our method is robust in a set-up with more than 2 agent groups in Appendix~\ref{app:robustness-semi}.

\subsection{Tuning the Region Size}%
\label{sub:semisynth_tuning_beta}

\begin{figure}[t]
\centering
  \begin{subfigure}[t]{0.47\textwidth}
  \centering
    \includegraphics[width=\linewidth]{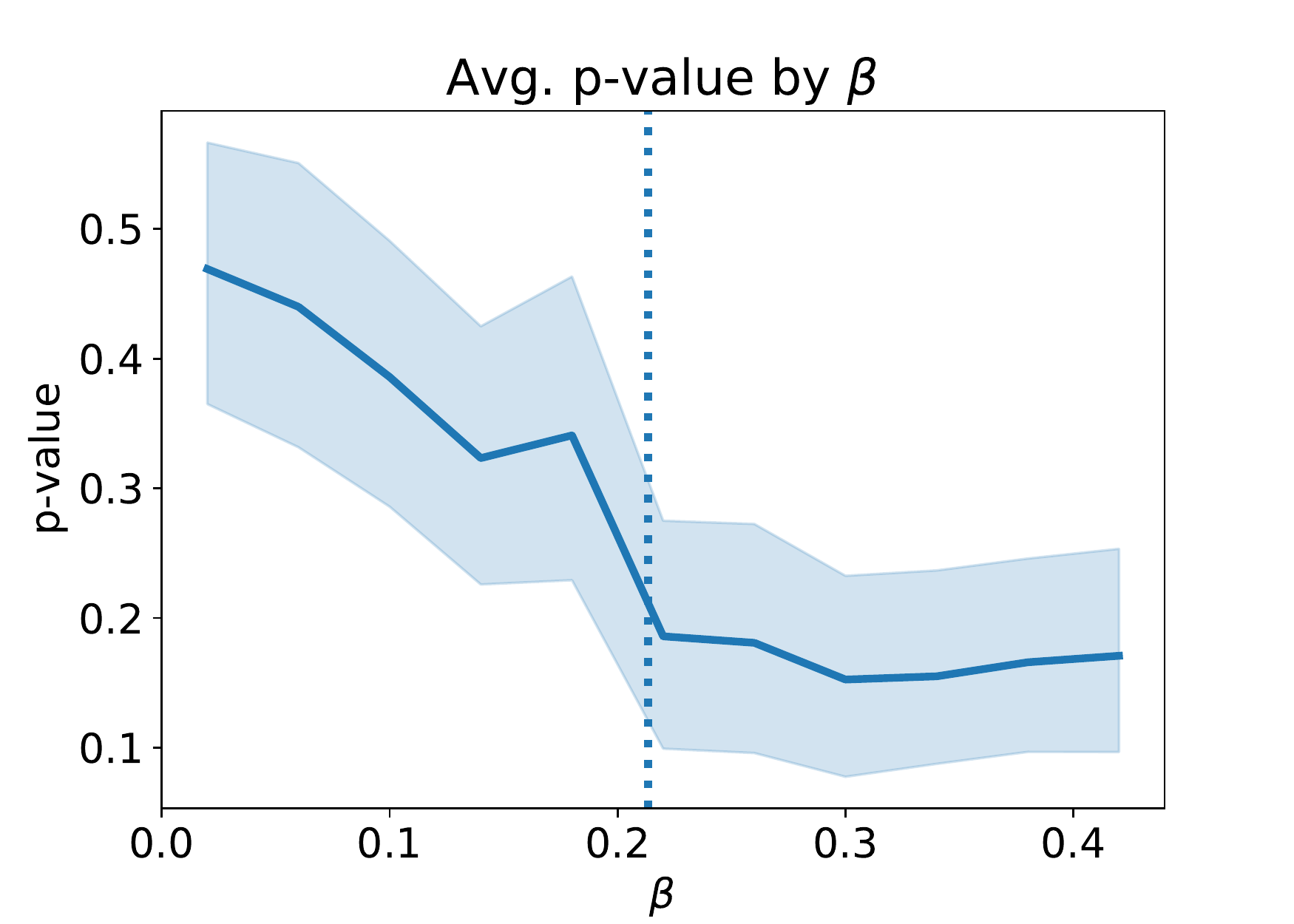}
  \caption{}%
  \label{fig:semisynth_beta_pvalue}
  \end{subfigure}
  \hspace{2em}
  \begin{subfigure}[t]{0.37\textwidth}
  \centering
     \includegraphics[width=\linewidth]{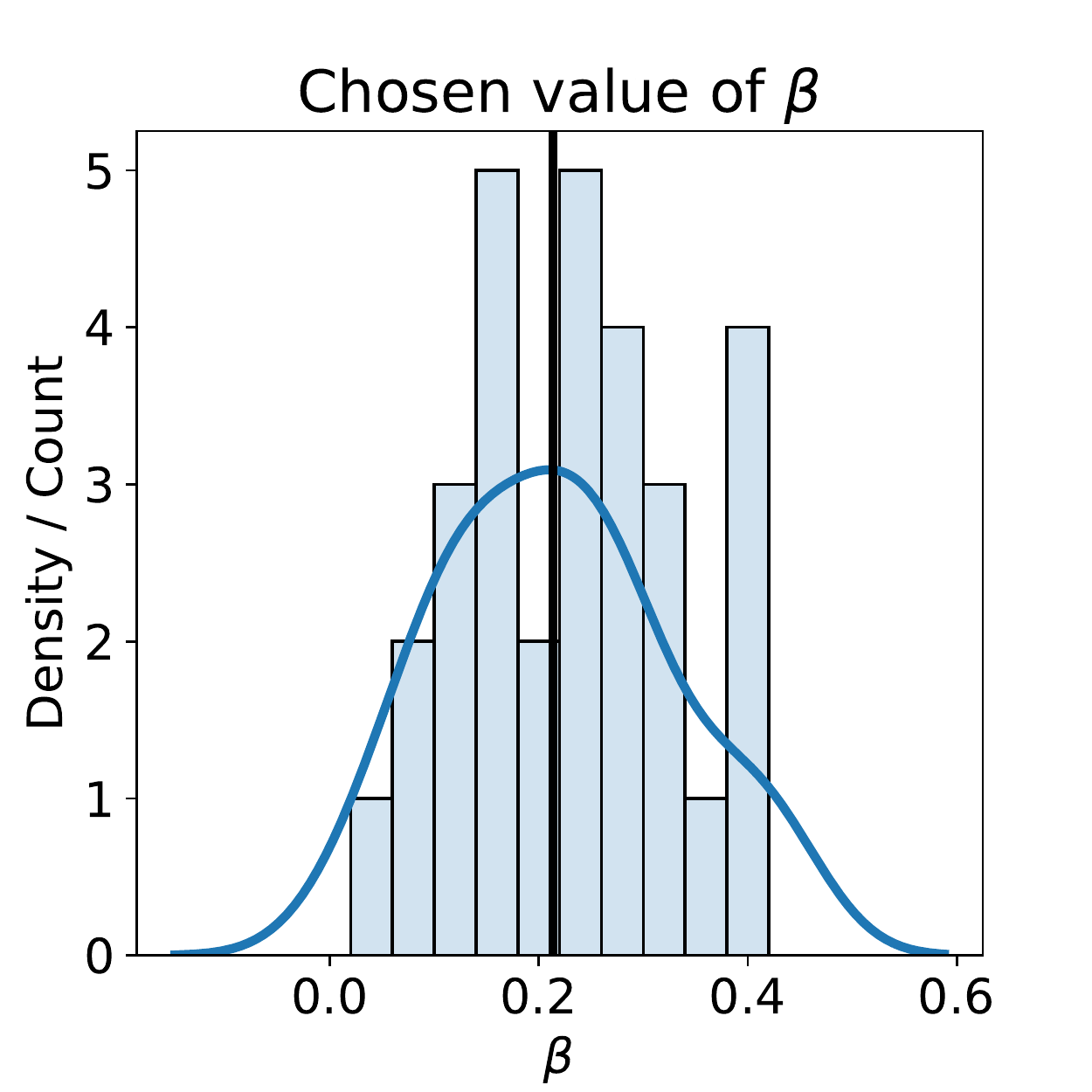}
  \caption{}%
  \label{fig:semisynth_beta_choices}
  \end{subfigure}
  \caption{Results of tuning $\beta$ over 30 semi-synthetic datasets. (\subref{fig:semisynth_beta_pvalue}) Average p-value for each candidate $\beta$, with 95\% uncertainty intervals for the mean generated by bootstrapping. The dotted vertical line represents the true value of $\beta$. (\subref{fig:semisynth_beta_choices}) Distribution of $\beta$ with the smallest p-value over the datasets.}%
\label{fig:semisynth_beta}
\end{figure}

We validate the proposed approach of tuning $\beta$ (discussed in Section~\ref{sub:tuning_the_region_size}), by applying the methodology to our semi-synthetic setting here.  We sample 30 semi-synthetic datasets and consider candidate values of $\beta$ in $[0.02, 0.42]$ in increments of $0.04$. For each proposed value of $\beta$ we use $T=40$ random permutations of the agents. Figure~\ref{fig:semisynth_beta} presents results for the misdemeanor under age 35 set-up.  As the candidate value of $\beta$ increases (up to the true value of $\beta$), the p-value decreases, and the distribution of selected $\beta$ values are centered on the true value of $\beta$.

\section{Real-data Experiment: First-Line Diabetes Treatment}%
\label{real-experiments}\label{sec:diabetes}

We apply our algorithm on a real-world dataset consisting of first-line (initial) treatment for type 2 diabetes and examine how the treatment variation we discover aligns with clinical knowledge.  We present an additional real-world experiment (on Parkinson's disease) in Appendix~\ref{sec:parkinsons}.

\textbf{Data and Setup}: We use an observational, de-identified, dataset provided by a large health insurer in the United States, spanning from 2010 to 2020. The task is to classify first-line treatment decisions  between metformin ($Y=0$)--the typical recommendation from the American Diabetes Association--and other common first-line treatments such as sitagliptin, glipizide, glimepiride, or glyburide ($Y=1$) \citep{american2010standards,hripcsak2016characterizing}.  As relevant clinical features, we include the patients' most recent eGFR (mL/min/1.73m2) and creatinine (mg/dL) measurements, incidence of heart failure, and treatment date. Because treatment date does not define a type of patient, we omit it from the region model. However, including it in the outcome model is essential because of increasing use of metformin over time. The agent indicator $A$ is the group practice of the doctor responsible for the patient's treatment. 3,980 patients and 447 group practices are included in our cohort. After requiring at least 4 patients per agent, 3,576 patients and 176 group practices are included. This filter ensures each group practice has at least 1 sample in the training and validation sets and at least 2 samples in the test set. In this experiment, we choose $\beta=0.25$ as input to our algorithm. See Appendix~\ref{sec:diabetes_app} for additional cohort definition and set-up details.

\textbf{Interpretation of Results}: To interpret the region, we use decision trees as our region model $h(x)$. The tree is visualized in Appendix~\ref{sec:diabetes_app}.  The decision tree identifies the region of heterogeneity as the union of \begin{enumerate*}[label=(\roman*)]
    \item eGFR below 71.5 and
    \item eGFR above 98.5 and creatinine above 0.815
\end{enumerate*}.
These regions align with clinical intuition. In the first region, low eGFR values indicate impaired renal function \citep{kidney2009kdigo}, which is a contraindication for metformin since it is traditionally associated with increased risk of lactic acidosis \citep{tahrani2007metformin}. Still, treatment decisions can vary here because guidelines for managing patients with eGFR below 45 are lacking \citep{guideline2015clinical}. Note that this region provides an example of how our algorithm works when overlap is not satisfied. Although 33 of 176 group practices do not see patients with these features, we can still conclude that this is a region of heterogeneity among the 143 agents with cases. In the second region, there are no obvious contraindications for metformin. Thus, understanding why some doctors on average only prescribe metformin 78\% of the time to patients in this region may help us identify whether this is a region in which we can standardize practice.

\textbf{Assessing Significance}: In Table~\ref{tab:IterativeAlgDecisionTree_benchmark_diabetes}, we perform a sanity check, assessing whether our algorithm identifies a region $S$ whose variation in held-out data is higher than that of a randomly selected region, using the partially optimized objective $L(S) = \max_{G} Q(S, G)$ laid out in Section~\ref{sub:hypothesis_test_validation} to compare regions directly. Table~\ref{tab:IterativeAlgDecisionTree_benchmark_diabetes} shows that $L(\hat{S})$ is similar on the training and test data. Furthermore, we compare  $L(\hat{S})$ on the test set to the distribution of $L(S_{\text{rand}})$, where $S_{\text{rand}}$ are random subsets of the test data of the same size as $\hat{S}$.  We compute the latter distribution using 100 random subsets and observe that the test value of $L(\hat{S})$ is more than two standard deviations above the mean of the latter.  This gives us confidence that the discovered region $S$ generalizes to a region of heterogeneity beyond the training set.
\begin{table}[t]
    \centering
    \caption{Objective values $L(S)$ for the learned region on the training and test datasets, along with the distribution of values for randomly generated regions $S_{\text{rand}}$ given as mean (standard deviation). }
        \begin{tabular}{lll}
        \toprule
        \textbf{Metric} & \textbf{Subset} & \textbf{Value} \\
        \midrule
         $L(\hat{S})$ & Train & 0.1029 \\
         $L(\hat{S})$ & Test & 0.0924 \\
         $L(S_{\text{rand}})$ & Test & 0.0507 (0.0073) \\
         \bottomrule
        \end{tabular}
    \label{tab:IterativeAlgDecisionTree_benchmark_diabetes}
\end{table}
We direct the reader to Appendix~\ref{sec:diabetes_app} for additional analyses, such as evaluating the stability of the region over multiple folds of splitting the data.

\section{Related Work}%
\label{related-work}

Beyond previously mentioned connections to causal effect estimation, we highlight a few areas of research that share similar goals to our own.

\textbf{Agent-specific Models of Decisions}: Prior works have modeled agent-specific decision-making processes by estimating a separate model for each agent. \citet{abaluck2016determinants} model heterogeneity in physician tendency to run diagnostic tests. \citet{chan2019selection} estimate radiologist skill based on diagnosis and miss rates. In our setting, unlike in diagnosis, there is no \enquote{correct} decision that can be incorporated into the model. \citet{ribers2020machine} assume there is provider-specific noise in determining patient type, which affects the pay-off functions for deciding whether to prescribe antibiotics. When only a few decisions are observed per agent, these agent-specific models cannot be estimated reliably.  \citet{currie2017diagnosing} also incorporate physician beliefs and skill into a pay-off function. They estimate an aggregate logistic choice model (for C-sections) across all physicians and then learn how individual physicians deviate from this model.  They do not learn the regions where this deviation occurs, as they focus on how heterogeneity is associated with downstream outcomes. \citet{norris2020examiner} looks for disagreement between agents but relies on some cases being seen by multiple agents. We assume each case is seen by only one decision-maker.

\textbf{Conditional Independence Testing}: While our objective maximizes a causal notion of dependence, one could instead ask if $Y$ is conditionally independent of $A$ given $X$. Many metrics exist for testing conditional independence, such as the Hilbert-Schmidt independence criterion (HSIC) \citep{fukumizu2007kernel, zhang2012kernel}, conditional mutual information \citep{runge2018conditional}, conditional correlation \citep{ramsey2014scalable}, and expected conditional covariance \citep{shah2020hardness}. We give the last a causal interpretation under some assumptions and seek a region that maximizes it, in lieu of testing.

\textbf{Hierarchical / Mixture Models}: Bayesian methods are often used to estimate group-level effects, such as a per-physician effect on patient outcomes \citep{Tuerk2008-no}, where group identifiers are included as a categorical feature in a multi-level generalized linear model \citep{Gelman2006-rw}. Alternatively, one could assume a conditional mixture model \citep{bishop2006pattern}, where agents belong to latent clusters that each have their own policy.  However, both of these methods require parametric assumptions on the distribution of $\P[Y \mid x, a]$, and even so, the optimal mixture model is not necessarily identifiable when both the clusters and policies are unknown \citep{dasgupta2007probabilistic}. By contrast, our method does not require making particular parametric assumptions about $\E[Y \mid x, a]$ and seeks to learn the region of heterogeneity directly. 

\textbf{Feature Evaluation}: Checking for heterogeneity can also be framed as feature evaluation, where we would like to evaluate whether adding the agent identifier will increase predictive power.  Feature evaluation methods typically maximize dependence between selected features and labels, utilizing measures similar to those in conditional independence testing, such as the HSIC \citep{song2007supervised}.  Other methods use the correlation of the new feature with the loss gradient of the current predictor as a measure of utility \citep{koepke2012fast}. In contrast, we focus not on checking marginally for the predictive power of agent identifiers, but rather identifying a region.

\textbf{Crowdsourcing}: Our work is conceptually related to identifying which samples are difficult to label via crowdsourcing annotations \citep{karger2014budget,whitehill2009whose}. Most crowdsourcing models are generative with latent variables for the correct sample labels, sample difficulty, agent expertise, and agent bias. They then optimize for the likelihood of the observed labels. The set of difficult samples is analogous to our region of heterogeneity. The main difference with our problem is that we do not require any notion of the \enquote{correct} label.

\section{Discussion}%
\label{sec:discussion}

In this work, we take a causal perspective on finding regions where agents (i.e., decision-makers) have heterogeneous preferences, formalizing this heterogeneity in terms of counterfactual (or \enquote{potential}) decisions across agents. We propose a causal measure of agent-specific biases and give an objective that aggregates this bias over agents.  This objective can be identified from observational data and written in terms of an expected conditional covariance.  Importantly, for our applications, this does not require building agent-specific models or assuming overlap across all agents.

We give an iterative algorithm to find a region that maximizes this objective. Then, we demonstrate in semi-synthetic experiments that our algorithm accurately recovers the region of heterogeneity and scales well with the number of agents.  In contrast, performance of baslines deteriorates when the number of agents increases.  Although our experiments have low-dimensional spaces, we hypothesize  our algorithm would scale well to high-dimensional feature spaces since the average policy and region models can handle high-dimensional input spaces. Finally, on a real-world medical dataset, we show that our algorithm can yield insights that align with clinical knowledge.

Our work is motivated by understanding variation in human decision-making.  In the judicial domain, our method may help unearth types of cases where decisions are highly dependent on the judge assigned to the case. In the medical domain, our approach may identify types of patients where new guidelines may be required to help doctors make decisions more consistent with standard of care. Domain expertise is required to determine the implications of the regions discovered by our  method. Beyond these applications, we see our approach as a useful data science tool for understanding heterogeneity in decisions that appears to be driven by individual-level preferences. 

\newpage
\textbf{Acknowledgements}: We would like to thank Aaron Smith-McLallen, James Denyer, Luogang Wei, Johnathon (Kyle) Armstrong, Neil Dixit, Aditya Sai, and the rest of the data science group at Independence Blue Cross, whose expertise, data, and support enabled the diabetes experiment. We would also like to thank Rebecca Boiarsky for converting the diabetes data to the OMOP common data model format, Monica Agrawal for her helpful comments, and other members of the lab for insightful discussions. We are also grateful to Charles Venuto and Monica Javidnia for their advice on Parkinson's.  This work was supported in part by Independence Blue Cross, Office of Naval Research Award No. N00014-17-1-2791, an Abdul Latif Jameel fellowship, and a NSF CAREER award.

\bibliographystyle{ACM-Reference-Format}
\bibliography{main}


\begin{thebibliography}{53}


\ifx \showCODEN    \undefined \def \showCODEN     #1{\unskip}     \fi
\ifx \showDOI      \undefined \def \showDOI       #1{#1}\fi
\ifx \showISBNx    \undefined \def \showISBNx     #1{\unskip}     \fi
\ifx \showISBNxiii \undefined \def \showISBNxiii  #1{\unskip}     \fi
\ifx \showISSN     \undefined \def \showISSN      #1{\unskip}     \fi
\ifx \showLCCN     \undefined \def \showLCCN      #1{\unskip}     \fi
\ifx \shownote     \undefined \def \shownote      #1{#1}          \fi
\ifx \showarticletitle \undefined \def \showarticletitle #1{#1}   \fi
\ifx \showURL      \undefined \def \showURL       {\relax}        \fi
\providecommand\bibfield[2]{#2}
\providecommand\bibinfo[2]{#2}
\providecommand\natexlab[1]{#1}
\providecommand\showeprint[2][]{arXiv:#2}

\bibitem[\protect\citeauthoryear{Abaluck, Agha, Kabrhel, Raja, and
  Venkatesh}{Abaluck et~al\mbox{.}}{2016}]%
        {abaluck2016determinants}
\bibfield{author}{\bibinfo{person}{Jason Abaluck}, \bibinfo{person}{Leila
  Agha}, \bibinfo{person}{Chris Kabrhel}, \bibinfo{person}{Ali Raja}, {and}
  \bibinfo{person}{Arjun Venkatesh}.} \bibinfo{year}{2016}\natexlab{}.
\newblock \showarticletitle{The determinants of productivity in medical
  testing: Intensity and allocation of care}.
\newblock \bibinfo{journal}{\emph{American Economic Review}}
  \bibinfo{volume}{106}, \bibinfo{number}{12} (\bibinfo{year}{2016}),
  \bibinfo{pages}{3730--64}.
\newblock


\bibitem[\protect\citeauthoryear{Aizer and Doyle~Jr}{Aizer and
  Doyle~Jr}{2013}]%
        {Aizer2013-ex}
\bibfield{author}{\bibinfo{person}{Anna Aizer} {and} \bibinfo{person}{Joseph~J
  Doyle~Jr}.} \bibinfo{year}{2013}\natexlab{}.
\newblock \bibinfo{booktitle}{\emph{Juvenile Incarceration, Human Capital and
  Future Crime: Evidence from Randomly-Assigned Judges}}.
\newblock \bibinfo{type}{{T}echnical {R}eport} w19102.
  \bibinfo{institution}{National Bureau of Economic Research}.
\newblock


\bibitem[\protect\citeauthoryear{{American Diabetes Association}}{{American
  Diabetes Association}}{2010}]%
        {american2010standards}
\bibfield{author}{\bibinfo{person}{{American Diabetes Association}}.}
  \bibinfo{year}{2010}\natexlab{}.
\newblock \showarticletitle{Standards of medical care in diabetes—2010}.
\newblock \bibinfo{journal}{\emph{Diabetes care}} \bibinfo{volume}{33},
  \bibinfo{number}{Supplement 1} (\bibinfo{year}{2010}),
  \bibinfo{pages}{S11--S61}.
\newblock


\bibitem[\protect\citeauthoryear{Arnold, Dobbie, and Yang}{Arnold
  et~al\mbox{.}}{2018}]%
        {Arnold2018-sw}
\bibfield{author}{\bibinfo{person}{David Arnold}, \bibinfo{person}{Will
  Dobbie}, {and} \bibinfo{person}{Crystal~S Yang}.}
  \bibinfo{year}{2018}\natexlab{}.
\newblock \showarticletitle{{Racial Bias in Bail Decisions}}.
\newblock \bibinfo{journal}{\emph{The Quarterly Journal of Economics}}
  \bibinfo{volume}{133}, \bibinfo{number}{4} (\bibinfo{date}{May}
  \bibinfo{year}{2018}), \bibinfo{pages}{1885--1932}.
\newblock


\bibitem[\protect\citeauthoryear{Atsma, Elwyn, and Westert}{Atsma
  et~al\mbox{.}}{2020}]%
        {Atsma2020}
\bibfield{author}{\bibinfo{person}{Femke Atsma}, \bibinfo{person}{Glyn Elwyn},
  {and} \bibinfo{person}{Gert Westert}.} \bibinfo{year}{2020}\natexlab{}.
\newblock \showarticletitle{{Understanding unwarranted variation in clinical
  practice: a focus on network effects, reflective medicine and learning health
  systems}}.
\newblock \bibinfo{journal}{\emph{International Journal for Quality in Health
  Care}} \bibinfo{volume}{32}, \bibinfo{number}{4} (\bibinfo{date}{jun}
  \bibinfo{year}{2020}), \bibinfo{pages}{271--274}.
\newblock
\showISSN{1353-4505}
\urldef\tempurl%
\url{https://doi.org/10.1093/intqhc/mzaa023}
\showDOI{\tempurl}


\bibitem[\protect\citeauthoryear{Birkmeyer, Reames, McCulloch, Carr, Campbell,
  and Wennberg}{Birkmeyer et~al\mbox{.}}{2013}]%
        {birkmeyer2013understanding}
\bibfield{author}{\bibinfo{person}{John~D Birkmeyer},
  \bibinfo{person}{Bradley~N Reames}, \bibinfo{person}{Peter McCulloch},
  \bibinfo{person}{Andrew~J Carr}, \bibinfo{person}{W~Bruce Campbell}, {and}
  \bibinfo{person}{John~E Wennberg}.} \bibinfo{year}{2013}\natexlab{}.
\newblock \showarticletitle{Understanding of regional variation in the use of
  surgery}.
\newblock \bibinfo{journal}{\emph{The Lancet}} \bibinfo{volume}{382},
  \bibinfo{number}{9898} (\bibinfo{year}{2013}), \bibinfo{pages}{1121--1129}.
\newblock


\bibitem[\protect\citeauthoryear{Bishop}{Bishop}{2006}]%
        {bishop2006pattern}
\bibfield{author}{\bibinfo{person}{Christopher~M Bishop}.}
  \bibinfo{year}{2006}\natexlab{}.
\newblock \bibinfo{booktitle}{\emph{Pattern recognition and machine learning}}.
\newblock \bibinfo{publisher}{springer}.
\newblock


\bibitem[\protect\citeauthoryear{Cabana, Rand, Powe, Wu, Wilson, Abboud, and
  Rubin}{Cabana et~al\mbox{.}}{1999}]%
        {cabana1999don}
\bibfield{author}{\bibinfo{person}{Michael~D Cabana},
  \bibinfo{person}{Cynthia~S Rand}, \bibinfo{person}{Neil~R Powe},
  \bibinfo{person}{Albert~W Wu}, \bibinfo{person}{Modena~H Wilson},
  \bibinfo{person}{Paul-Andr{\'e}~C Abboud}, {and} \bibinfo{person}{Haya~R
  Rubin}.} \bibinfo{year}{1999}\natexlab{}.
\newblock \showarticletitle{Why don't physicians follow clinical practice
  guidelines?: A framework for improvement}.
\newblock \bibinfo{journal}{\emph{Jama}} \bibinfo{volume}{282},
  \bibinfo{number}{15} (\bibinfo{year}{1999}), \bibinfo{pages}{1458--1465}.
\newblock


\bibitem[\protect\citeauthoryear{Chan~Jr, Gentzkow, and Yu}{Chan~Jr
  et~al\mbox{.}}{2019}]%
        {chan2019selection}
\bibfield{author}{\bibinfo{person}{David~C Chan~Jr}, \bibinfo{person}{Matthew
  Gentzkow}, {and} \bibinfo{person}{Chuan Yu}.}
  \bibinfo{year}{2019}\natexlab{}.
\newblock \bibinfo{booktitle}{\emph{Selection with variation in diagnostic
  skill: Evidence from radiologists}}.
\newblock \bibinfo{type}{{T}echnical {R}eport}. \bibinfo{institution}{National
  Bureau of Economic Research}.
\newblock


\bibitem[\protect\citeauthoryear{Corallo, Croxford, Goodman, Bryan, Srivastava,
  and Stukel}{Corallo et~al\mbox{.}}{2014}]%
        {corallo2014systematic}
\bibfield{author}{\bibinfo{person}{Ashley~N Corallo}, \bibinfo{person}{Ruth
  Croxford}, \bibinfo{person}{David~C Goodman}, \bibinfo{person}{Elisabeth~L
  Bryan}, \bibinfo{person}{Divya Srivastava}, {and} \bibinfo{person}{Therese~A
  Stukel}.} \bibinfo{year}{2014}\natexlab{}.
\newblock \showarticletitle{A systematic review of medical practice variation
  in OECD countries}.
\newblock \bibinfo{journal}{\emph{Health Policy}} \bibinfo{volume}{114},
  \bibinfo{number}{1} (\bibinfo{year}{2014}), \bibinfo{pages}{5--14}.
\newblock


\bibitem[\protect\citeauthoryear{Currie and MacLeod}{Currie and
  MacLeod}{2017}]%
        {currie2017diagnosing}
\bibfield{author}{\bibinfo{person}{Janet Currie} {and}
  \bibinfo{person}{W~Bentley MacLeod}.} \bibinfo{year}{2017}\natexlab{}.
\newblock \showarticletitle{Diagnosing expertise: Human capital, decision
  making, and performance among physicians}.
\newblock \bibinfo{journal}{\emph{Journal of labor economics}}
  \bibinfo{volume}{35}, \bibinfo{number}{1} (\bibinfo{year}{2017}),
  \bibinfo{pages}{1--43}.
\newblock


\bibitem[\protect\citeauthoryear{Dasgupta and Schulman}{Dasgupta and
  Schulman}{2007}]%
        {dasgupta2007probabilistic}
\bibfield{author}{\bibinfo{person}{Sanjoy Dasgupta} {and}
  \bibinfo{person}{Leonard~J Schulman}.} \bibinfo{year}{2007}\natexlab{}.
\newblock \showarticletitle{A probabilistic analysis of EM for mixtures of
  separated, spherical Gaussians}.
\newblock \bibinfo{journal}{\emph{Journal of Machine Learning Research}}
  \bibinfo{volume}{8} (\bibinfo{year}{2007}), \bibinfo{pages}{203--226}.
\newblock


\bibitem[\protect\citeauthoryear{De~Jong, Westert, Lagoe, and
  Groenewegen}{De~Jong et~al\mbox{.}}{2006}]%
        {de2006variation}
\bibfield{author}{\bibinfo{person}{Judith~D De~Jong}, \bibinfo{person}{Gert~P
  Westert}, \bibinfo{person}{Ronald Lagoe}, {and} \bibinfo{person}{Peter~P
  Groenewegen}.} \bibinfo{year}{2006}\natexlab{}.
\newblock \showarticletitle{Variation in hospital length of stay: do physicians
  adapt their length of stay decisions to what is usual in the hospital where
  they work?}
\newblock \bibinfo{journal}{\emph{Health Services Research}}
  \bibinfo{volume}{41}, \bibinfo{number}{2} (\bibinfo{year}{2006}),
  \bibinfo{pages}{374--394}.
\newblock


\bibitem[\protect\citeauthoryear{Di~Tella and Schargrodsky}{Di~Tella and
  Schargrodsky}{2013}]%
        {Di_Tella2013-sp}
\bibfield{author}{\bibinfo{person}{Rafael Di~Tella} {and}
  \bibinfo{person}{Ernesto Schargrodsky}.} \bibinfo{year}{2013}\natexlab{}.
\newblock \showarticletitle{Criminal Recidivism after Prison and Electronic
  Monitoring}.
\newblock \bibinfo{journal}{\emph{The Journal of Political Economy}}
  \bibinfo{volume}{121}, \bibinfo{number}{1} (\bibinfo{year}{2013}),
  \bibinfo{pages}{28--73}.
\newblock


\bibitem[\protect\citeauthoryear{Dressel and Farid}{Dressel and Farid}{2018}]%
        {Dressel2018-br}
\bibfield{author}{\bibinfo{person}{Julia Dressel} {and} \bibinfo{person}{Hany
  Farid}.} \bibinfo{year}{2018}\natexlab{}.
\newblock \showarticletitle{{The accuracy, fairness, and limits of predicting
  recidivism}}.
\newblock \bibinfo{journal}{\emph{Science advances}} \bibinfo{volume}{4},
  \bibinfo{number}{1} (\bibinfo{date}{Jan.} \bibinfo{year}{2018}),
  \bibinfo{pages}{eaao5580}.
\newblock


\bibitem[\protect\citeauthoryear{Fukumizu, Gretton, Sun, and
  Sch{\"o}lkopf}{Fukumizu et~al\mbox{.}}{2007}]%
        {fukumizu2007kernel}
\bibfield{author}{\bibinfo{person}{Kenji Fukumizu}, \bibinfo{person}{Arthur
  Gretton}, \bibinfo{person}{Xiaohai Sun}, {and} \bibinfo{person}{Bernhard
  Sch{\"o}lkopf}.} \bibinfo{year}{2007}\natexlab{}.
\newblock \showarticletitle{Kernel measures of conditional dependence.}. In
  \bibinfo{booktitle}{\emph{NeurIPS}}, Vol.~\bibinfo{volume}{20}.
  \bibinfo{pages}{489--496}.
\newblock


\bibitem[\protect\citeauthoryear{Gelman and Hill}{Gelman and Hill}{2006}]%
        {Gelman2006-rw}
\bibfield{author}{\bibinfo{person}{Andrew Gelman} {and}
  \bibinfo{person}{Jennifer Hill}.} \bibinfo{year}{2006}\natexlab{}.
\newblock \bibinfo{booktitle}{\emph{Data Analysis Using Regression and
  Multilevel/Hierarchical Models}}.
\newblock \bibinfo{publisher}{Cambridge University Press}.
\newblock


\bibitem[\protect\citeauthoryear{Goetz, Fahn, Martinez-Martin, Poewe, Sampaio,
  Stebbins, Stern, Tilley, Dodel, Dubois, et~al\mbox{.}}{Goetz
  et~al\mbox{.}}{2007}]%
        {goetz2007movement}
\bibfield{author}{\bibinfo{person}{Christopher~G Goetz},
  \bibinfo{person}{Stanley Fahn}, \bibinfo{person}{Pablo Martinez-Martin},
  \bibinfo{person}{Werner Poewe}, \bibinfo{person}{Cristina Sampaio},
  \bibinfo{person}{Glenn~T Stebbins}, \bibinfo{person}{Matthew~B Stern},
  \bibinfo{person}{Barbara~C Tilley}, \bibinfo{person}{Richard Dodel},
  \bibinfo{person}{Bruno Dubois}, {et~al\mbox{.}}}
  \bibinfo{year}{2007}\natexlab{}.
\newblock \showarticletitle{Movement Disorder Society-sponsored revision of the
  Unified Parkinson's Disease Rating Scale (MDS-UPDRS): process, format, and
  clinimetric testing plan}.
\newblock \bibinfo{journal}{\emph{Movement disorders}} \bibinfo{volume}{22},
  \bibinfo{number}{1} (\bibinfo{year}{2007}), \bibinfo{pages}{41--47}.
\newblock


\bibitem[\protect\citeauthoryear{Group, Bilo, Coentr{\~a}o, Couchoud, Covic,
  De~Sutter, Drechsler, Gnudi, Goldsmith, Heaf, et~al\mbox{.}}{Group
  et~al\mbox{.}}{2015}]%
        {guideline2015clinical}
\bibfield{author}{\bibinfo{person}{Guideline~Development Group},
  \bibinfo{person}{Henk Bilo}, \bibinfo{person}{Luis Coentr{\~a}o},
  \bibinfo{person}{C{\'e}cile Couchoud}, \bibinfo{person}{Adrian Covic},
  \bibinfo{person}{Johan De~Sutter}, \bibinfo{person}{Christiane Drechsler},
  \bibinfo{person}{Luigi Gnudi}, \bibinfo{person}{David Goldsmith},
  \bibinfo{person}{James Heaf}, {et~al\mbox{.}}}
  \bibinfo{year}{2015}\natexlab{}.
\newblock \showarticletitle{Clinical practice guideline on management of
  patients with diabetes and chronic kidney disease stage 3b or higher (eGFR<
  45 mL/min)}.
\newblock \bibinfo{journal}{\emph{Nephrology Dialysis Transplantation}}
  \bibinfo{volume}{30}, \bibinfo{number}{suppl\_2} (\bibinfo{year}{2015}),
  \bibinfo{pages}{ii1--ii142}.
\newblock


\bibitem[\protect\citeauthoryear{Group et~al\mbox{.}}{Group
  et~al\mbox{.}}{2009}]%
        {kidney2009kdigo}
\bibfield{author}{\bibinfo{person}{Kidney Disease: Improving Global Outcomes
  (KDIGO) CKD-MBD~Work Group} {et~al\mbox{.}}} \bibinfo{year}{2009}\natexlab{}.
\newblock \showarticletitle{KDIGO clinical practice guideline for the
  diagnosis, evaluation, prevention, and treatment of Chronic Kidney
  Disease-Mineral and Bone Disorder (CKD-MBD)}.
\newblock \bibinfo{journal}{\emph{Kidney international. Supplement}}
  \bibinfo{number}{113} (\bibinfo{year}{2009}), \bibinfo{pages}{S1--S130}.
\newblock


\bibitem[\protect\citeauthoryear{Group et~al\mbox{.}}{Group
  et~al\mbox{.}}{2014}]%
        {pd2014long}
\bibfield{author}{\bibinfo{person}{PD~Med~Collaborative Group} {et~al\mbox{.}}}
  \bibinfo{year}{2014}\natexlab{}.
\newblock \showarticletitle{Long-term effectiveness of dopamine agonists and
  monoamine oxidase B inhibitors compared with levodopa as initial treatment
  for Parkinson's disease (PD MED): a large, open-label, pragmatic randomised
  trial}.
\newblock \bibinfo{journal}{\emph{The Lancet}} \bibinfo{volume}{384},
  \bibinfo{number}{9949} (\bibinfo{year}{2014}), \bibinfo{pages}{1196--1205}.
\newblock


\bibitem[\protect\citeauthoryear{Hern{\'a}n and Robins}{Hern{\'a}n and
  Robins}{2020}]%
        {Hernan2020-zj}
\bibfield{author}{\bibinfo{person}{M~A Hern{\'a}n} {and} \bibinfo{person}{J~M
  Robins}.} \bibinfo{year}{2020}\natexlab{}.
\newblock \bibinfo{booktitle}{\emph{{Causal Inference: What If}}}.
\newblock \bibinfo{publisher}{Chapman \& Hall/CRC}, \bibinfo{address}{Boca
  Raton}.
\newblock


\bibitem[\protect\citeauthoryear{Hripcsak, Ryan, Duke, Shah, Park, Huser,
  Suchard, Schuemie, DeFalco, Perotte, et~al\mbox{.}}{Hripcsak
  et~al\mbox{.}}{2016}]%
        {hripcsak2016characterizing}
\bibfield{author}{\bibinfo{person}{George Hripcsak}, \bibinfo{person}{Patrick~B
  Ryan}, \bibinfo{person}{Jon~D Duke}, \bibinfo{person}{Nigam~H Shah},
  \bibinfo{person}{Rae~Woong Park}, \bibinfo{person}{Vojtech Huser},
  \bibinfo{person}{Marc~A Suchard}, \bibinfo{person}{Martijn~J Schuemie},
  \bibinfo{person}{Frank~J DeFalco}, \bibinfo{person}{Adler Perotte},
  {et~al\mbox{.}}} \bibinfo{year}{2016}\natexlab{}.
\newblock \showarticletitle{Characterizing treatment pathways at scale using
  the OHDSI network}.
\newblock \bibinfo{journal}{\emph{Proceedings of the National Academy of
  Sciences}} \bibinfo{volume}{113}, \bibinfo{number}{27}
  (\bibinfo{year}{2016}), \bibinfo{pages}{7329--7336}.
\newblock


\bibitem[\protect\citeauthoryear{Kang, Bennett, Carbado, Casey, Dasgupta,
  Faigman, Godsil, Greenwald, Levinson, and Mnookin}{Kang
  et~al\mbox{.}}{2012}]%
        {Kang2012}
\bibfield{author}{\bibinfo{person}{Jerry Kang}, \bibinfo{person}{Judge~Mark
  Bennett}, \bibinfo{person}{Devon Carbado}, \bibinfo{person}{Pam Casey},
  \bibinfo{person}{Nilanjana Dasgupta}, \bibinfo{person}{David Faigman},
  \bibinfo{person}{Rachel Godsil}, \bibinfo{person}{Anthony~G. Greenwald},
  \bibinfo{person}{Justin Levinson}, {and} \bibinfo{person}{Jennifer Mnookin}.}
  \bibinfo{year}{2012}\natexlab{}.
\newblock \showarticletitle{{Implicit bias in the courtroom}}.
\newblock \bibinfo{journal}{\emph{UCLA Law Review}} \bibinfo{volume}{59},
  \bibinfo{number}{5} (\bibinfo{year}{2012}), \bibinfo{pages}{1124--1186}.
\newblock


\bibitem[\protect\citeauthoryear{Karger, Oh, and Shah}{Karger
  et~al\mbox{.}}{2014}]%
        {karger2014budget}
\bibfield{author}{\bibinfo{person}{David~R Karger}, \bibinfo{person}{Sewoong
  Oh}, {and} \bibinfo{person}{Devavrat Shah}.} \bibinfo{year}{2014}\natexlab{}.
\newblock \showarticletitle{Budget-optimal task allocation for reliable
  crowdsourcing systems}.
\newblock \bibinfo{journal}{\emph{Operations Research}} \bibinfo{volume}{62},
  \bibinfo{number}{1} (\bibinfo{year}{2014}), \bibinfo{pages}{1--24}.
\newblock


\bibitem[\protect\citeauthoryear{Kleinberg, Lakkaraju, Leskovec, Ludwig, and
  Mullainathan}{Kleinberg et~al\mbox{.}}{2018}]%
        {Kleinberg2018-ni}
\bibfield{author}{\bibinfo{person}{Jon Kleinberg}, \bibinfo{person}{Himabindu
  Lakkaraju}, \bibinfo{person}{Jure Leskovec}, \bibinfo{person}{Jens Ludwig},
  {and} \bibinfo{person}{Sendhil Mullainathan}.}
  \bibinfo{year}{2018}\natexlab{}.
\newblock \showarticletitle{{Human Decisions and Machine Predictions}}.
\newblock \bibinfo{journal}{\emph{The Quarterly Journal of Economics}}
  \bibinfo{volume}{133}, \bibinfo{number}{1} (\bibinfo{date}{Feb.}
  \bibinfo{year}{2018}), \bibinfo{pages}{237--293}.
\newblock


\bibitem[\protect\citeauthoryear{Kling}{Kling}{2006}]%
        {Kling2006-qm}
\bibfield{author}{\bibinfo{person}{Jeffrey~R Kling}.}
  \bibinfo{year}{2006}\natexlab{}.
\newblock \showarticletitle{{Incarceration Length, Employment, and Earnings}}.
\newblock \bibinfo{journal}{\emph{The American economic review}}
  \bibinfo{volume}{96}, \bibinfo{number}{3} (\bibinfo{date}{June}
  \bibinfo{year}{2006}), \bibinfo{pages}{863--876}.
\newblock


\bibitem[\protect\citeauthoryear{Koepke and Bilenko}{Koepke and
  Bilenko}{2012}]%
        {koepke2012fast}
\bibfield{author}{\bibinfo{person}{Hoyt Koepke} {and} \bibinfo{person}{Mikhail
  Bilenko}.} \bibinfo{year}{2012}\natexlab{}.
\newblock \showarticletitle{Fast prediction of new feature utility}.
\newblock \bibinfo{journal}{\emph{arXiv preprint arXiv:1206.4680}}
  (\bibinfo{year}{2012}).
\newblock


\bibitem[\protect\citeauthoryear{Lin, Jung, Goel, and Skeem}{Lin
  et~al\mbox{.}}{2020}]%
        {Lin2020-ev}
\bibfield{author}{\bibinfo{person}{Zhiyuan~(Jerry) Lin},
  \bibinfo{person}{Jongbin Jung}, \bibinfo{person}{Sharad Goel}, {and}
  \bibinfo{person}{Jennifer Skeem}.} \bibinfo{year}{2020}\natexlab{}.
\newblock \showarticletitle{The Limits of Human Predictions of Recidivism}.
\newblock \bibinfo{journal}{\emph{Science Advances}} \bibinfo{volume}{6},
  \bibinfo{number}{7} (\bibinfo{year}{2020}).
\newblock
\urldef\tempurl%
\url{https://doi.org/10.1126/sciadv.aaz0652}
\showDOI{\tempurl}
\showeprint{https://advances.sciencemag.org/content/6/7/eaaz0652.full.pdf}


\bibitem[\protect\citeauthoryear{Marek, Jennings, Lasch, Siderowf, Tanner,
  Simuni, Coffey, Kieburtz, Flagg, Chowdhury, et~al\mbox{.}}{Marek
  et~al\mbox{.}}{2011}]%
        {marek2011parkinson}
\bibfield{author}{\bibinfo{person}{Kenneth Marek}, \bibinfo{person}{Danna
  Jennings}, \bibinfo{person}{Shirley Lasch}, \bibinfo{person}{Andrew
  Siderowf}, \bibinfo{person}{Caroline Tanner}, \bibinfo{person}{Tanya Simuni},
  \bibinfo{person}{Chris Coffey}, \bibinfo{person}{Karl Kieburtz},
  \bibinfo{person}{Emily Flagg}, \bibinfo{person}{Sohini Chowdhury},
  {et~al\mbox{.}}} \bibinfo{year}{2011}\natexlab{}.
\newblock \showarticletitle{The parkinson progression marker initiative
  (PPMI)}.
\newblock \bibinfo{journal}{\emph{Progress in neurobiology}}
  \bibinfo{volume}{95}, \bibinfo{number}{4} (\bibinfo{year}{2011}),
  \bibinfo{pages}{629--635}.
\newblock


\bibitem[\protect\citeauthoryear{Muzerengi and Clarke}{Muzerengi and
  Clarke}{2015}]%
        {muzerengi2015initial}
\bibfield{author}{\bibinfo{person}{Sharon Muzerengi} {and}
  \bibinfo{person}{Carl~E Clarke}.} \bibinfo{year}{2015}\natexlab{}.
\newblock \showarticletitle{Initial drug treatment in Parkinson’s disease}.
\newblock \bibinfo{journal}{\emph{bmj}}  \bibinfo{volume}{351}
  (\bibinfo{year}{2015}).
\newblock


\bibitem[\protect\citeauthoryear{Nie and Wager}{Nie and Wager}{2017}]%
        {nie2017quasi}
\bibfield{author}{\bibinfo{person}{Xinkun Nie} {and} \bibinfo{person}{Stefan
  Wager}.} \bibinfo{year}{2017}\natexlab{}.
\newblock \showarticletitle{Quasi-oracle estimation of heterogeneous treatment
  effects}.
\newblock \bibinfo{journal}{\emph{arXiv preprint arXiv:1712.04912}}
  (\bibinfo{year}{2017}).
\newblock


\bibitem[\protect\citeauthoryear{Norris}{Norris}{2020}]%
        {norris2020examiner}
\bibfield{author}{\bibinfo{person}{S Norris}.} \bibinfo{year}{2020}\natexlab{}.
\newblock \bibinfo{booktitle}{\emph{Examiner inconsistency: Evidence from
  refugee decisions}}.
\newblock \bibinfo{type}{{T}echnical {R}eport}. \bibinfo{institution}{Working
  paper}.
\newblock


\bibitem[\protect\citeauthoryear{Ortega, Wang, Wu, and Olynk}{Ortega
  et~al\mbox{.}}{2011}]%
        {Ortega2011}
\bibfield{author}{\bibinfo{person}{David~L Ortega}, \bibinfo{person}{H~Holly
  Wang}, \bibinfo{person}{Laping Wu}, {and} \bibinfo{person}{Nicole~J Olynk}.}
  \bibinfo{year}{2011}\natexlab{}.
\newblock \showarticletitle{{Modeling heterogeneity in consumer preferences for
  select food safety attributes in China}}.
\newblock \bibinfo{journal}{\emph{Food Policy}} \bibinfo{volume}{36},
  \bibinfo{number}{2} (\bibinfo{year}{2011}), \bibinfo{pages}{318--324}.
\newblock
\urldef\tempurl%
\url{https://www.sciencedirect.com/science/article/pii/S0306919210001442}
\showURL{%
\tempurl}


\bibitem[\protect\citeauthoryear{Paszke, Gross, Massa, Lerer, Bradbury, Chanan,
  Killeen, Lin, Gimelshein, Antiga, et~al\mbox{.}}{Paszke
  et~al\mbox{.}}{2019}]%
        {paszke2019pytorch}
\bibfield{author}{\bibinfo{person}{Adam Paszke}, \bibinfo{person}{Sam Gross},
  \bibinfo{person}{Francisco Massa}, \bibinfo{person}{Adam Lerer},
  \bibinfo{person}{James Bradbury}, \bibinfo{person}{Gregory Chanan},
  \bibinfo{person}{Trevor Killeen}, \bibinfo{person}{Zeming Lin},
  \bibinfo{person}{Natalia Gimelshein}, \bibinfo{person}{Luca Antiga},
  {et~al\mbox{.}}} \bibinfo{year}{2019}\natexlab{}.
\newblock \showarticletitle{Pytorch: An imperative style, high-performance deep
  learning library}.
\newblock \bibinfo{journal}{\emph{arXiv preprint arXiv:1912.01703}}
  (\bibinfo{year}{2019}).
\newblock


\bibitem[\protect\citeauthoryear{Pearl}{Pearl}{2009}]%
        {Pearl2009-sj}
\bibfield{author}{\bibinfo{person}{Judea Pearl}.}
  \bibinfo{year}{2009}\natexlab{}.
\newblock \bibinfo{booktitle}{\emph{{{Causality: Models, Reasoning, and
  Inference}}} (\bibinfo{edition}{2nd} ed.)}.
\newblock \bibinfo{publisher}{Cambridge University Press}.
\newblock


\bibitem[\protect\citeauthoryear{Pedregosa, Varoquaux, Gramfort, Michel,
  Thirion, Grisel, Blondel, Prettenhofer, Weiss, Dubourg, Vanderplas, Passos,
  Cournapeau, Brucher, Perrot, and Duchesnay}{Pedregosa et~al\mbox{.}}{2011}]%
        {scikit-learn}
\bibfield{author}{\bibinfo{person}{F. Pedregosa}, \bibinfo{person}{G.
  Varoquaux}, \bibinfo{person}{A. Gramfort}, \bibinfo{person}{V. Michel},
  \bibinfo{person}{B. Thirion}, \bibinfo{person}{O. Grisel},
  \bibinfo{person}{M. Blondel}, \bibinfo{person}{P. Prettenhofer},
  \bibinfo{person}{R. Weiss}, \bibinfo{person}{V. Dubourg}, \bibinfo{person}{J.
  Vanderplas}, \bibinfo{person}{A. Passos}, \bibinfo{person}{D. Cournapeau},
  \bibinfo{person}{M. Brucher}, \bibinfo{person}{M. Perrot}, {and}
  \bibinfo{person}{E. Duchesnay}.} \bibinfo{year}{2011}\natexlab{}.
\newblock \showarticletitle{Scikit-learn: Machine Learning in {P}ython}.
\newblock \bibinfo{journal}{\emph{Journal of Machine Learning Research}}
  \bibinfo{volume}{12} (\bibinfo{year}{2011}), \bibinfo{pages}{2825--2830}.
\newblock


\bibitem[\protect\citeauthoryear{Ramsey}{Ramsey}{2014}]%
        {ramsey2014scalable}
\bibfield{author}{\bibinfo{person}{Joseph~D Ramsey}.}
  \bibinfo{year}{2014}\natexlab{}.
\newblock \showarticletitle{A scalable conditional independence test for
  nonlinear, non-gaussian data}.
\newblock \bibinfo{journal}{\emph{arXiv preprint arXiv:1401.5031}}
  (\bibinfo{year}{2014}).
\newblock


\bibitem[\protect\citeauthoryear{Research}{Research}{2019}]%
        {econml}
\bibfield{author}{\bibinfo{person}{Microsoft Research}.}
  \bibinfo{year}{2019}\natexlab{}.
\newblock \bibinfo{title}{{EconML}: {A Python Package for ML-Based
  Heterogeneous Treatment Effects Estimation}}.
\newblock \bibinfo{howpublished}{https://github.com/microsoft/EconML}.
\newblock
\newblock
\shownote{Version 0.x.}


\bibitem[\protect\citeauthoryear{Ribers and Ullrich}{Ribers and
  Ullrich}{2020}]%
        {ribers2020machine}
\bibfield{author}{\bibinfo{person}{Michael~A Ribers} {and}
  \bibinfo{person}{Hannes Ullrich}.} \bibinfo{year}{2020}\natexlab{}.
\newblock \showarticletitle{Machine Predictions and Human Decisions with
  Variation in Payoffs and Skill}.
\newblock  (\bibinfo{year}{2020}).
\newblock


\bibitem[\protect\citeauthoryear{Runge}{Runge}{2018}]%
        {runge2018conditional}
\bibfield{author}{\bibinfo{person}{Jakob Runge}.}
  \bibinfo{year}{2018}\natexlab{}.
\newblock \showarticletitle{Conditional independence testing based on a
  nearest-neighbor estimator of conditional mutual information}. In
  \bibinfo{booktitle}{\emph{International Conference on Artificial Intelligence
  and Statistics}}. PMLR, \bibinfo{pages}{938--947}.
\newblock


\bibitem[\protect\citeauthoryear{Scarpa, Philippidis, and Spalatro}{Scarpa
  et~al\mbox{.}}{2005}]%
        {Scarpa2005}
\bibfield{author}{\bibinfo{person}{Riccardo Scarpa}, \bibinfo{person}{George
  Philippidis}, {and} \bibinfo{person}{Fiorenza Spalatro}.}
  \bibinfo{year}{2005}\natexlab{}.
\newblock \showarticletitle{{Product-country images and preference
  heterogeneity for Mediterranean food products: A discrete choice framework}}.
\newblock \bibinfo{journal}{\emph{Agribusiness}} \bibinfo{volume}{21},
  \bibinfo{number}{3} (\bibinfo{year}{2005}), \bibinfo{pages}{329--349}.
\newblock


\bibitem[\protect\citeauthoryear{Shah and Peters}{Shah and Peters}{2020}]%
        {shah2020hardness}
\bibfield{author}{\bibinfo{person}{Rajen~D. Shah} {and} \bibinfo{person}{Jonas
  Peters}.} \bibinfo{year}{2020}\natexlab{}.
\newblock \bibinfo{title}{The Hardness of Conditional Independence Testing and
  the Generalised Covariance Measure}.
\newblock
\newblock
\showeprint[arxiv]{1804.07203}~[math.ST]


\bibitem[\protect\citeauthoryear{Shalev-Shwartz and Ben-David}{Shalev-Shwartz
  and Ben-David}{2014}]%
        {Shalev-Shwartz2014}
\bibfield{author}{\bibinfo{person}{Shai Shalev-Shwartz} {and}
  \bibinfo{person}{Shai Ben-David}.} \bibinfo{year}{2014}\natexlab{}.
\newblock \bibinfo{booktitle}{\emph{Understanding Machine Learning: From Theory
  to Algorithms}}. Vol.~\bibinfo{volume}{9781107057}.
\newblock \bibinfo{publisher}{Cambridge University Press}.
\newblock
\showISBNx{9781107298019}
\urldef\tempurl%
\url{https://doi.org/10.1017/CBO9781107298019}
\showDOI{\tempurl}


\bibitem[\protect\citeauthoryear{Shalit, Johansson, and Sontag}{Shalit
  et~al\mbox{.}}{2017}]%
        {shalit2017estimating}
\bibfield{author}{\bibinfo{person}{Uri Shalit}, \bibinfo{person}{Fredrik~D
  Johansson}, {and} \bibinfo{person}{David Sontag}.}
  \bibinfo{year}{2017}\natexlab{}.
\newblock \showarticletitle{Estimating individual treatment effect:
  generalization bounds and algorithms}. In
  \bibinfo{booktitle}{\emph{International Conference on Machine Learning}}.
  PMLR, \bibinfo{pages}{3076--3085}.
\newblock


\bibitem[\protect\citeauthoryear{Song, Smola, Gretton, Borgwardt, and
  Bedo}{Song et~al\mbox{.}}{2007}]%
        {song2007supervised}
\bibfield{author}{\bibinfo{person}{Le Song}, \bibinfo{person}{Alex Smola},
  \bibinfo{person}{Arthur Gretton}, \bibinfo{person}{Karsten~M Borgwardt},
  {and} \bibinfo{person}{Justin Bedo}.} \bibinfo{year}{2007}\natexlab{}.
\newblock \showarticletitle{Supervised feature selection via dependence
  estimation}. In \bibinfo{booktitle}{\emph{Proceedings of the 24th
  international conference on Machine learning}}. \bibinfo{pages}{823--830}.
\newblock


\bibitem[\protect\citeauthoryear{Tahrani, Varughese, Scarpello, and
  Hanna}{Tahrani et~al\mbox{.}}{2007}]%
        {tahrani2007metformin}
\bibfield{author}{\bibinfo{person}{AA Tahrani}, \bibinfo{person}{GI Varughese},
  \bibinfo{person}{JH Scarpello}, {and} \bibinfo{person}{FWF Hanna}.}
  \bibinfo{year}{2007}\natexlab{}.
\newblock \showarticletitle{Metformin, heart failure, and lactic acidosis: is
  metformin absolutely contraindicated?}
\newblock \bibinfo{journal}{\emph{Bmj}} \bibinfo{volume}{335},
  \bibinfo{number}{7618} (\bibinfo{year}{2007}), \bibinfo{pages}{508--512}.
\newblock


\bibitem[\protect\citeauthoryear{Tuerk, Mueller, and Egede}{Tuerk
  et~al\mbox{.}}{2008}]%
        {Tuerk2008-no}
\bibfield{author}{\bibinfo{person}{Peter~W Tuerk}, \bibinfo{person}{Martina
  Mueller}, {and} \bibinfo{person}{Leonard~E Egede}.}
  \bibinfo{year}{2008}\natexlab{}.
\newblock \showarticletitle{Estimating physician effects on glycemic control in
  the treatment of diabetes: methods, effects sizes, and implications for
  treatment policy}.
\newblock \bibinfo{journal}{\emph{Diabetes care}} \bibinfo{volume}{31},
  \bibinfo{number}{5} (\bibinfo{date}{May} \bibinfo{year}{2008}),
  \bibinfo{pages}{869--873}.
\newblock


\bibitem[\protect\citeauthoryear{Wager and Athey}{Wager and Athey}{2018}]%
        {wager2018estimation}
\bibfield{author}{\bibinfo{person}{Stefan Wager} {and} \bibinfo{person}{Susan
  Athey}.} \bibinfo{year}{2018}\natexlab{}.
\newblock \showarticletitle{Estimation and inference of heterogeneous treatment
  effects using random forests}.
\newblock \bibinfo{journal}{\emph{J. Amer. Statist. Assoc.}}
  \bibinfo{volume}{113}, \bibinfo{number}{523} (\bibinfo{year}{2018}),
  \bibinfo{pages}{1228--1242}.
\newblock


\bibitem[\protect\citeauthoryear{Waskom}{Waskom}{2021}]%
        {Waskom2021seaborn}
\bibfield{author}{\bibinfo{person}{Michael~L. Waskom}.}
  \bibinfo{year}{2021}\natexlab{}.
\newblock \showarticletitle{seaborn: statistical data visualization}.
\newblock \bibinfo{journal}{\emph{Journal of Open Source Software}}
  \bibinfo{volume}{6}, \bibinfo{number}{60} (\bibinfo{year}{2021}),
  \bibinfo{pages}{3021}.
\newblock
\urldef\tempurl%
\url{https://doi.org/10.21105/joss.03021}
\showDOI{\tempurl}


\bibitem[\protect\citeauthoryear{Wasserman}{Wasserman}{2004}]%
        {Wasserman2004-qm}
\bibfield{author}{\bibinfo{person}{Larry Wasserman}.}
  \bibinfo{year}{2004}\natexlab{}.
\newblock \bibinfo{booktitle}{\emph{{All of Statistics: A Concise Course in
  Statistical Inference}}}.
\newblock \bibinfo{publisher}{Springer, New York, NY}.
\newblock


\bibitem[\protect\citeauthoryear{Whitehill, Wu, Bergsma, Movellan, and
  Ruvolo}{Whitehill et~al\mbox{.}}{2009}]%
        {whitehill2009whose}
\bibfield{author}{\bibinfo{person}{Jacob Whitehill}, \bibinfo{person}{Ting-fan
  Wu}, \bibinfo{person}{Jacob Bergsma}, \bibinfo{person}{Javier Movellan},
  {and} \bibinfo{person}{Paul Ruvolo}.} \bibinfo{year}{2009}\natexlab{}.
\newblock \showarticletitle{Whose vote should count more: Optimal integration
  of labels from labelers of unknown expertise}.
\newblock \bibinfo{journal}{\emph{Advances in neural information processing
  systems}}  \bibinfo{volume}{22} (\bibinfo{year}{2009}),
  \bibinfo{pages}{2035--2043}.
\newblock


\bibitem[\protect\citeauthoryear{Zhang, Peters, Janzing, and
  Sch{\"o}lkopf}{Zhang et~al\mbox{.}}{2012}]%
        {zhang2012kernel}
\bibfield{author}{\bibinfo{person}{Kun Zhang}, \bibinfo{person}{Jonas Peters},
  \bibinfo{person}{Dominik Janzing}, {and} \bibinfo{person}{Bernhard
  Sch{\"o}lkopf}.} \bibinfo{year}{2012}\natexlab{}.
\newblock \showarticletitle{Kernel-based conditional independence test and
  application in causal discovery}.
\newblock \bibinfo{journal}{\emph{arXiv preprint arXiv:1202.3775}}
  (\bibinfo{year}{2012}).
\newblock


\end{thebibliography}

\clearpage 
\appendix

\section*{Appendix}%
\label{sec:appendix}

This appendix contains the following sections:
\begin{itemize}
    \item \textbf{Proofs}: Section~\ref{sec:proofs} contains proofs of theoretical results presented in the main paper.
    \item \textbf{Semi-Synthetic Experiment: Additional Details}: Section~\ref{sec:semi-synth-details-app} contains additional details and experimental results for the semi-synthetic experiment presented in Section~\ref{sec:semi_synthetic_experiments}.
    \item \textbf{Diabetes Experiment: Additional Details}: Section~\ref{sec:diabetes_app} contains additional details and experimental results for the real-data diabetes experiment presented in Section~\ref{sec:diabetes}.
    \item \textbf{Additional Real-Data Experiment: Parkinson's}: Section~\ref{sec:extra_real_data_experiments_app} contains an additional real-data experiment, on a medical dataset of patients with Parkinson's disease.
\end{itemize}
Most of our experiments were run on CPUs, with only the TARNet baseline run on a GEForce GTX GPU. We estimate the compute time to be on the order of 100 hours.

\section{Proofs of Theoretical Results}
\label{sec:proofs}

\subsection{Proof of Minor Claims in Section~\ref{sec:causal_perspective}}
\label{sec:proof_ypi}

\begin{thmappprop}\label{prop:ypi_conditional_exp}
Under the assumptions of consistency and no-unmeasured-confounding (NUC), the following equivalence holds
\[Y(\pi(x)) = \E[Y \mid x],\]
where $Y(\pi(x))$ is defined as $\sum_{a'} \E[Y(a') \mid X = x] \pi(a' \mid x)$, where $\pi(a' \mid x) \coloneqq \P(A = a' \mid X = x)$.
\end{thmappprop}
\begin{proof}
Based on the definition of $Y(\pi(x))$, we can write it as follows, using the fact that $Y$ is binary.
\begin{align*}
   Y(\pi(x)) &\coloneqq \sum_{a} \P(Y(a') = 1 \mid X = x) \pi(a' \mid x) \\
   &= \sum_{a'} \P(Y(a') = 1 \mid A = a', X = x) \P(A = a' \mid X = x) & \text{(NUC)} \\
   &= \sum_{a'} \P(Y = 1 \mid A = a', X = x) \P(A = a' \mid X = x) & \text{(Consistency)} \\
   &= \sum_{a'} \P(Y = 1, A = a' \mid X = x) \\
   &= \E[Y \mid X = x]
\end{align*}
\end{proof}

\begin{thmappprop}\label{prop:rewrite_conditional_relative_agent_bias}
Under the assumptions of consistency and no-unmeasured-confounding (NUC), the following equivalence holds
\begin{equation*}
\E[Y(a) - Y(\pi) \mid A = a, X \in S] = \int_{x \in S} \E[Y(a) - Y(\pi) \mid x] p(x \mid A = a, X \in S) dx,
\end{equation*}
\end{thmappprop}
\begin{proof}
This can be seen as follows, noting that by our definition of $Y(\pi(x))$ as a function of $x$, $\E[Y(\pi(x)) \mid X = x] = Y(\pi(x))$
\begin{align*}
\E[Y(a) - Y(\pi) \mid A = a, X \in S] &= \int_{x \in S} \E[(Y(a) - \E[Y \mid x]) \mid A = a, X = x] p(x \mid a, X \in S) dx \\
&= \int_{x \in S} \left(\E[Y(a) \mid X = x]  - \E[Y \mid x] \right) p(x \mid a, X \in S) dx & \text{(NUC)} \\
&= \int_{x \in S} \E[Y(a) - Y(\pi) \mid x] p(x \mid A = a, X \in S) dx
\end{align*}
\end{proof}

\begin{thmappprop}\label{prop:crab_as_residuals}
Under the assumptions of consistency and no-unmeasured-confounding (NUC), the conditional relative agent bias can be written as 
\begin{equation*}
\E[Y(a) - Y(\pi) \mid A = a, X \in S] = \E[Y - \E[Y \mid X] \mid A = a, X \in S]
\end{equation*}
\end{thmappprop}
\begin{proof}
By Propositions~\ref{prop:ypi_conditional_exp} and~\ref{prop:rewrite_conditional_relative_agent_bias}, we can re-write the conditional relative agent bias as 
\begin{align*}
&\E[Y(a) - Y(\pi) \mid A = a, X \in S] \\
= &\int_{x \in S} \E[Y(a) - Y(\pi) \mid x] p(x \mid A = a, X \in S) dx & \text{Prop.~\ref{prop:rewrite_conditional_relative_agent_bias}} \\
= &\int_{x \in S} \E[Y(a) - \E[Y \mid x] \mid x] p(x \mid A = a, X \in S) dx & \text{Prop.~\ref{prop:ypi_conditional_exp}} \\
= &\int_{x \in S} \E[Y(a) \mid x] p(x \mid A = a, X \in S) dx \\
&\quad - \int_{x \in S} \E[\E[Y \mid x] \mid x] p(x \mid A = a, X \in S) dx \\
= &\int_{x \in S} \E[Y \mid X = x, A = a] p(x \mid A = a, X \in S) dx \\
&\quad - \int_{x \in S} \E[\E[Y \mid x] \mid x] p(x \mid A = a, X \in S) dx & \text{(NUC)} \\
= &\int_{x \in S} \E[Y \mid X = x, A = a, X \in S] p(x \mid A = a, X \in S) dx \\
&\quad - \int_{x \in S} \E[\E[Y \mid x] \mid X = x, A = a, X \in S] p(x \mid A = a, X \in S) dx \\
= &\int_{x \in S} \E[Y - \E[Y \mid x] \mid X = x, A = a, X \in S] p(x \mid A = a, X \in S) dx \\
= &\E[Y - \E[Y \mid x] \mid A = a, X \in S],
\end{align*}
where the third-to-last line follows from the fact that the event $\{X = x \land A = a\} \iff \{X = x \land A = a \land X \in S\}$ over the set of $x$ that we are integrating over, and thus does not change the conditional expectation of $Y$.  Meanwhile, $\E[Y \mid x]$ is a function of $x$ alone, and so the conditional expectation is equivalent if we condition on additional information $\E[ \E[Y \mid x] \mid x] = \E[ \E[Y \mid x] \mid X = x, A = a, X \in S]$ as long as this conditional expectation is well-defined, which it will be wherever $p(x \mid A = a, X \in S) > 0$.
\end{proof}

\begin{thmappprop}\label{prop:equivalence_to_abs_values}
The partially maximized population objective from Equation~\eqref{eq:rewriting_abs_values} is equivalent (up to a factor of 2) to a weighted sum of the absolute value of each agent's conditional relative agent bias.  In other words:
\begin{align*}
  &\sum_{a \in \cA} \P(A = a \mid X \in S) \abs{\E[Y - \E[Y\mid X] \mid A = a, X \in S]}_{+} \\
  &\quad \quad = \frac{1}{2} \sum_{a \in \cA} \P(A = a \mid X \in S) \abs{\E[Y - \E[Y\mid X] \mid A = a, X \in S]}
\end{align*}
\end{thmappprop}
\begin{proof}
\begin{align*}
  &\E[Y - \E[Y \mid X] \mid X \in S] \\
  &= \sum_{a \in \cA} \P(A = a \mid X \in S) \E[Y - \E[Y\mid X] \mid A = a, X \in S] \\
  &= \sum_{a \in \cA} \P(A = a \mid X \in S) \left(\abs{\E[Y - \E[Y\mid X] \mid A = a, X \in S]}_{+} + \abs{\E[Y - \E[Y\mid X] \mid A = a, X \in S]}_{-} \right) \\
  &= 0 
\end{align*}
where $\abs{x}_{-} = \min(x, 0)$ denotes the negative part, and where the last line follows from the fact that $\E[Y - \E[Y \mid X] \mid X \in S] = 0$, from the definition of the conditional expectation.  This implies that 
\begin{align*}
  &\sum_{a \in \cA} \P(A = a \mid X \in S) \abs{\E[Y - \E[Y\mid X] \mid A = a, X \in S]}_{+} \\
  &= - \sum_{a \in \cA} \P(A = a \mid X \in S) \abs{\E[Y - \E[Y\mid X] \mid A = a, X \in S]}_{-},
\end{align*}
while the weighted sum of the absolute values is given by 
\begin{align*}
  &\sum_{a \in \cA} \P(A = a \mid X \in S) \abs{\E[Y - \E[Y\mid X] \mid A = a, X \in S]} \\
  &=\sum_{a \in \cA} \P(A = a \mid X \in S) \left(\abs{\E[Y - \E[Y\mid X] \mid A = a, X \in S]}_{+} - \abs{\E[Y - \E[Y\mid X] \mid A = a, X \in S]}_{-} \right) \\
  &=2 \sum_{a \in \cA} \P(A = a \mid X \in S) \abs{\E[Y - \E[Y\mid X] \mid A = a, X \in S]}_{+} 
\end{align*}
where the last line follows from the fact that the weighted sum of the positive parts is equal to the (negative) weighted sum of the negative parts.
\end{proof}

\subsection{Proof of Theorem~\ref{thmthm:causal_objective}}%
\label{sub:proof_of_theorem_thmthm}

\begin{thmappthm}[Causal Identification]\label{thmthm:causal_objective_app}
Under Assumption~\ref{thmasmp:causal_identification}, $Q(S, G)$ can be identified as 
\begin{align}
  Q(S, G) = \E_S[\cov(Y, G \mid X)] = \E_S[(Y - \E[Y \mid X]) G]
\end{align}
where $\E_{S}[\cdot] \coloneqq \E[\cdot \mid X \in S]$, and $\cov(Y, G \mid X)$ is the conditional covariance.
\end{thmappthm}

First, we will prove the following lemma: 

\begin{thmlem}\label{thmlem:causal_id_a}
Under Assumption~\ref{thmasmp:causal_identification}, we can write the expected conditional covariance as follows for binary random variables $Y, \1{A = a}$
\begin{align*}
  \E[\cov(Y, \1{A = a} \mid x) \mid X \in S] &= \P(A = a \mid X \in S) \E[Y(a) - Y(\pi(x)) \mid A = a, X \in S]
\end{align*}
\end{thmlem}
\begin{proof}
The conditional covariance can be written as follows
\begin{align*}
\cov(Y, \1{A = a} \mid x) &= \E[(Y - \E[Y \mid x]) \1{A = a} \mid x] \\
&=\P(Y = 1, A = a \mid x) - \P(Y = 1 \mid x) \P(A = a \mid x) \\
&=(\E[Y \mid A = a, x] - \E[Y \mid x]) \P(A = a \mid x) \\
&=(\E[Y(a) \mid x] - \E[Y \mid x]) \P(A = a \mid x) \\
&= \E[Y(a) - Y \mid x] \P(A = a \mid x)
\end{align*}
where in the penultimate line, we use our causal assumptions to write that \[\E[Y \mid A = a, x] = \E[Y(a) \mid A = a, x] = \E[Y(a) \mid x],\] by consistency and no-unmeasured-confounding, respectively. To get the expected conditional covariance, we integrate this over $x \in S$ to arrive at 
\begin{align*}
\E[\cov(Y, A \mid x) \mid X \in S] &= \int_{x \in S} \E[Y(a) - Y \mid x] \P(A = a \mid x) p(x \mid X \in S) dx \\
&= \int_{x \in S} \E[Y(a) - Y \mid x] \P(A = a \mid x, X \in S) p(x \mid X \in S) dx \\ 
&= \int_{x \in S} \E[Y(a) - Y \mid x] \P(A = a, X = x \mid X \in S) \\ 
&= \P(A = a \mid X \in S) \int_{x \in S} \E[Y(a) - Y \mid x] p(x \mid A = a, X \in S) \\  
&= \P(A = a \mid X \in S) \E[Y(a) - \E[Y \mid x]] \mid A = a, X \in S]
\end{align*}
where in the second line, we note that $\P(A = a \mid x) = \P(A = a \mid x, X \in S)$, since the event $\{X = x\} \subset \{X \in S\}$ for any $x \in S$.
\end{proof}

With this in hand, we can prove Theorem~\ref{thmthm:causal_objective} by noting that we can write the function $G$ as $G(A) = \sum_{a: G(a) = 1} \1{A = a}$.  Using this, we can write that 
\begin{align*}
  \E[\cov(Y, G(A) \mid x) \mid X \in S] &= \E[\cov(Y, \sum_{a: G(a) = 1} \1{A = a} \mid x) \mid X \in S] \\
                           &= \E[\sum_{a: G(a) = 1} \cov(Y, \1{A = a} \mid x) \mid X \in S] \\
                           &= \sum_{a: G(a) = 1} \E[\cov(Y, \1{A = a} \mid x) \mid X \in S] \\
                           &= \sum_{a: G(a) = 1} \P(A = a \mid X \in S) \E[Y(a) - Y(\pi(x)) \mid A = a, X \in S]
\end{align*}
where the second equality follows from linearity of the conditional covariance, the third line follows from linearity of expectation, and the last line follows from Lemma~\ref{thmlem:causal_id_a}

\subsection{Proof of Covariance Identity}%
\label{app:cov-id}
In the main text, we claimed that $\E[\cov(U, V\mid X)]=\E[(U-\E[U\mid X])V]$ for binary $U, V$.  This is a known fact, but we give a short a proof here for completeness.
\begin{proof}
Let $U, V$ be binary random variables, then 
\begin{align}
    &\cov(U, V\mid X)=\E[(U-\E[U\mid X])(V-\E[V\mid X])\mid X] \nonumber \\
    &=\E[(U-\E[U\mid X])V\mid X]-\E[(U-\E[U\mid X])\E[V\mid X]\mid X] \nonumber \\
    &=\E[(U-\E[U\mid X])V\mid X]. \label{lem1_fact}
\end{align}
Here, Eq.~\eqref{lem1_fact} follows since for any bounded $f(X)$,
\begin{align*}
    \E[(U-\E[U\mid X])f(X)\mid X]&=f(X)\E[U-\E[U\mid X]\mid X]=0.
\end{align*}
The result follows from taking expectation with respect to $X$.
\end{proof}

\subsection{Proof that $S', G'$ Maximizes Objective \ref{eq:objective}}
\label{app:s_maximizes_objective}
\begin{proof}
For any $S$ and $G$, we have:
\begin{align*}
    Q(S, G)&=\E_S[(Y-\E[Y\mid X])G]\\
    &=\E_S\left[(Y-\E[Y\mid X])\left(\sum_{a} G(a)\1{A=a} \right)\right]\\
    &=\sum_{a}G(a)\E_S[(Y-\E[Y\mid X])\1{A=a}]\\
    &=\sum_{a}G(a)\E_S[Y-\E[Y\mid X]\mid A=a]\P(A=a\mid X\in S).
\end{align*}
This quantity is maximized by picking $G(a)=1$ if $\E_S[Y-\E[Y\mid X]\mid A=a]\ge 0$ and $G(a)=0$ otherwise. If $S$ were disjoint from $S'$, $\E_S[Y-\E[Y\mid X]\mid A=a]=0$ by Assumption \ref{assumption:zero_outside_s}. If $S$ intersects $S'$, Assumption \ref{assumption:separable} implies that $\E_S[Y-\E[Y\mid X]\mid A=a]$ is positive if $G'(a)=1$ and negative otherwise. In both cases, $G'$ maximizes $Q(S, G)$ for a fixed $S$, i.e. $Q(S, G)\le Q(S, G')$. By Assumptions \ref{assumption:zero_outside_s} and \ref{assumption:separable}, we know that $\cov(Y, G'\mid X=x)>0$ for all $x\in S'$, and $\cov(Y, G'\mid X)=0$ for all $x\notin S'$. Since $\P(S')=\beta$ and Objective \ref{eq:objective} requires $\P(S)\ge \beta$, the optimal choice must be to take $S=S'$. Therefore, $Q(S', G')$ must maximize our objective, as desired, which justifies our writing them as $S^*, G^*$.
\end{proof}

\subsection{Proof of Proposition~\ref{thmprop:optimal_partition}}
\label{app:thm1-proof}
\begin{proof}
Notice that 
\begin{align*}
\hat{Q}(S, G) &=\frac{1}{\sum_{a, j}\1{x_{aj}\in S}}\sum_{a, j} (y_{aj}-f(x_{aj}))\cdot G(a) \cdot \1{x_{aj} \in S}\\
&=\sum_{a} G(a) \frac{1}{\sum_{a, j}\1{x_{aj}\in S}}\sum_{j} (y_{aj}-f(x_{aj})) \cdot \1{x_{aj} \in S}\\
&=\sum_{a} G(a)\hat{Q}(S, \1{A=a}).
\end{align*}
To maximize this quantity, the optimal choice is $G_S$, where $G_S(a)=1$ if $\hat{Q}(S, \1{A=a})\ge 0$, and $G_S(a)=0$ otherwise. To minimize this quantity, the opposite choice suffices, giving us $G_S^c$.
\end{proof}

\subsection{Proof of Theorem \ref{thm:correctness}}
\label{app:thm2-proof}

We first present the full version of the Theorem, without the simplifications.  Let $\beta'$ denote the fraction of our $N\cdot R$ samples that fall in the region $S^*$.  This is the version of the Theorem that we will prove:
\begin{thmappthm}[Theorem 2, Formal]
Under the same assumptions as in Section~\ref{sec:algorithm_analysis}, as long as $S^*$ is in our hypothesis class $\cS$ and $R > \frac{2 \ln 2}{\alpha^2 \beta^2 \omega^2}$, the first iteration of Algorithm~\ref{alg:main} returns $\hat{S}$ such that, with probability at least $1-\delta$, $Q(S^*, G^*)-Q(\hat{S}, G^*)\le \epsilon$, where
\begin{align*}
   \epsilon &= \sqrt{\frac{2\ln(3/\delta)}{\beta'N\cdot R}}+\left(\frac{1}{\beta'}+\frac{1}{\beta}\right)\left(\eta+\sqrt{\frac{3\eta(1-\eta)}{\delta \cdot N}}\right) \\
   &\quad\quad +\abs{\frac{1}{\hat{\P}(\hat{S})} - \frac{1}{\P(\hat{S})}} + \frac{1}{\P(\hat{S})}\left( 2 \cR(\cS, N\cdot R) + 4 \sqrt{\frac{2 \ln (12 / \delta)}{N \cdot R}} \right),
\end{align*}
where $\cR(\cS, N \cdot R)$ is the Rademacher complexity of $\cS$, and
\[
    \eta=\exp\left(\frac{-R\alpha^2\beta^2\omega^2}{2}\right).
\]
\end{thmappthm}

For simplicity, we assume that $X$ is a continuous random variable with a well defined density $\P(X = x)$.  We first bound the error rate in estimating the grouping on the first iteration.
\begin{thmlem}\label{thmlem:noise_rate}
Under Assumptions~\ref{assumption:homogeneity},~\ref{assumption:separable}, and~\ref{assumption:coverage}, let $\hat{G}$ be the grouping returned by the first iteration of Algorithm~\ref{alg:main}. Then for every $a \in A$, 
\begin{align}
\P[\hat{G}(a) \neq G^*(a)] &\leq \eta, & \text{where } && \eta &\coloneqq \exp\left(\frac{-R\alpha^2\beta^2\omega^2}{2}\right). \label{eq:def_eta}
\end{align}
with $\alpha$ defined in Assumption~\ref{assumption:separable}, $\beta$ given as input to the algorithm, and $\omega$ defined in Assumption~\ref{assumption:coverage}. As long as $R > \frac{2\ln 2}{\alpha^2 \beta^2 \omega^2}$, then $\eta < 1/2$
\end{thmlem}
\begin{proof}
Choose some agent $a$, and assume that $G^*(a) = 1$, as the argument is symmetric if $G^*(a) = 0$. Define $\hat{Q}_a \coloneqq \frac{1}{R}\sum_{j=1}^R (y_{aj}-f(x_{aj}))$ as shorthand for the sample average $\hat{Q}(S, \1{A = a})$, where $S = \cX$ for the first iteration. Then 
\begin{equation*}
\hat{G}(a) = \1{Q_a \geq 0},
\end{equation*}
and since $G^*(a) = 1$, it suffices to bound the probability that $Q_a < 0$.  The expected value of $Q_a$ is given by 
\begin{align}
    &\E[Q_a] = \E[Y-\E[Y\mid X]\mid A=a] \nonumber\\
    &=\E\left[(Y-\E[Y\mid X]) \1{X \in S^*} + (Y-\E[Y\mid X]) \1{X \not\in S^*} \mid A=a\right] \label{eq:using_assumption_zero_outside_s}\\
    &=\E[(Y-\E[Y\mid X]) \1{X \in S^*} \mid A=a] \nonumber \\
    &=\int_{\tilde{S}} \E[Y-\E[Y\mid X]\mid X=x, A=a]\P(X=x\mid A=a)\,dx & \tilde{S} \coloneqq \{x: x \in S^* \land p(x \mid a) > 0\} \nonumber\\
    &\ge \int_{\tilde{S}}\alpha \P(X = x \mid A = a) \,dx\label{eq:using_assumption_separable}\\
    &= \alpha \P(X \in S^* \mid A = a)  \\
    &\geq \alpha\beta\omega, \label{eq:using_assumption_coverage}
\end{align}
where the second term in Eq. \eqref{eq:using_assumption_zero_outside_s} is zero by Assumption \ref{assumption:zero_outside_s}; in the inequality on Eq. \eqref{eq:using_assumption_separable} we have used Assumptions \ref{assumption:homogeneity} and \ref{assumption:separable} to lower-bound $\E[Y - \E[Y \mid X] \mid X = x, A = a]$ (recall that $G^*(a) = 1$), and in Eq. \eqref{eq:using_assumption_coverage} we have used Assumption \ref{assumption:coverage} to lower bound $\P(X \in S^* \mid A = a)$ and we have used the definition of $\beta$ as $\P(X \in S^*)$.

Because $f(x)=\E[Y\mid X]$ by assumption, $\hat{Q}_{a}$ is an average of i.i.d samples $(y_{aj}-f(x_{aj}))$ drawn from $\P(Y-\E[Y\mid X]\mid A=a)$. Since $y_{aj}\in \{0, 1\}$ and $f(x_{i, j})\in [0, 1]$, each sample is bounded by the interval $[-1, 1]$. By Hoeffding's inequality, the probability of misclassification is bounded as
\begin{align*}
    \P(\hat{Q}_a\le 0) &=\P(\E[Q_a]-\hat{Q}_a\ge \E[Q_a])) \\
    &\le \exp\left(\frac{-R\alpha^2\beta^2\omega^2}{2}\right) =: \eta.
\end{align*}
since $\E[Q_a] \geq \alpha \beta \omega$. A symmetric argument holds for $G^*(a)=0$.
\end{proof}
\begin{proof}[Proof of Theorem~\ref{thm:correctness}]
After the first iteration, Algorithm~\ref{alg:main} returns a grouping $\hat{G}$ and a subset $\hat{S}$.  We are interested in the quality of this set $\hat{S}$, relative to the optimal set $S^*$.  For an agent $a$, let $\hat{g}_a \coloneqq \hat{G}(a)$, and let $g_a \coloneqq G^*(a)$.  For notational convenience, let $r_{aj} \coloneqq y_{aj} - f(x_{aj})$ be the residual in predicting the treatment. We now have $NR$ samples of the form $r_{aj} \cdot \hat{g}_a$. We let $\hat{Q}$ be the empirical expected conditional covariance computed from samples, so that
\begin{align}
\hat{Q}(S, G^*) &=\frac{1}{\sum_{a, j}\1{x_{aj}\in S}}\sum_{a, j} r_{aj} \cdot g_a \cdot \1{x_{aj} \in S},
\end{align}
and similarly for $\hat{Q}(S, \hat{G})$. Let $\hat{S}=\argmax_{\cS}\hat{Q}(S, \hat{G})$, then we can expand by adding and subtracting identical terms
\begin{align}
    &Q(S^*, G^*)-Q(\hat{S}, G^*) \nonumber\\
    =&[Q(S^*, G^*)-\hat{Q}(S^*, \hat{G})]+[\hat{Q}(S^*, \hat{G})-\hat{Q}(\hat{S}, \hat{G})] + [\hat{Q}(\hat{S}, \hat{G})-Q(\hat{S}, G^*)]\nonumber\\
    \le& [Q(S^*, G^*)-\hat{Q}(S^*, \hat{G})]+[\hat{Q}(\hat{S}, \hat{G})-Q(\hat{S}, G^*)] \label{eq:empirical_minimized}\\
    = &[Q(S^*, G^*)-\hat{Q}(S^*, G^*)]+[\hat{Q}(S^*, G^*)-\hat{Q}(S^*, \hat{G})]+\nonumber\\
    &\qquad [\hat{Q}(\hat{S}, \hat{G})-\hat{Q}(\hat{S}, G^*)]+[\hat{Q}(\hat{S}, G^*)-Q(\hat{S}, G^*)], \label{eq:erm_decomposition}
\end{align}
where in Eq. \eqref{eq:empirical_minimized} we have used the fact that $\hat{S}$ is the maximizer of $\hat{Q}(S, \hat{G})$ in our hypothesis class, and the assumption that $S^*$ is in our hypothesis class.  We will bound these terms in order. 

\textbf{Bounding the first term of Eq.~\eqref{eq:erm_decomposition}:} For any $S$, let $N_S$ be the number of samples $x_{aj}\in S$. Since $\abs{r_{aj} \cdot g_{a}} \leq 1$, we have by Hoeffding's inequality that for any $\epsilon_0>0$,
\begin{equation}
\label{eq:concentration_bound}
   \P(Q(S^*, G^*)-\hat{Q}(S^*, G^*)>\epsilon_0)\le\exp\left(-\frac{N_{S^*}\epsilon_0^2}{2}\right)=\exp\left(-\frac{\beta'NR\epsilon_0^2}{2}\right),
\end{equation}
where we have used the fact that $N_{S^*}=\beta'NR$, by definition of $\beta'$.

\textbf{Bounding the second, third terms of Eq.~\eqref{eq:erm_decomposition}}: For any $S$, 
\begin{align}
    \hat{Q}(S, G^*)-\hat{Q}(S, \hat{G})&=\frac{1}{N_S}\sum_{{a, j}} r_{aj} \cdot (g_a-\hat{g}_a) \cdot \1{x_{aj} \in S} \nonumber \\
    &\le \frac{1}{N_S}\sum_{{a, j}} \abs{r_{aj}} \cdot \abs{(g_a-\hat{g}_a)} \cdot \1{x_{aj} \in S} \nonumber \\
    &\le \frac{1}{N_S}\sum_{a, j} \abs{g_a-\hat{g}_a} =\frac{RN}{N_S}\left(\frac{1}{N}\sum_a|g_a-\hat{g}_a|\right), \label{eq:bound_errors}
\end{align}
where for simplicity, there are $R$ samples per agent by assumption.  Note that $\abs{g_a-\hat{g_a}}$ is 1 if $g_a$ is misclassified, and 0 otherwise.  Each $\hat{g}_a$ is independently distributed, and by Lemma~\ref{thmlem:noise_rate}, $p_a \coloneqq \P(g_{a} \neq \hat{g}_a) \leq \eta$.  Then $\E[\sum_{a} \abs{g_a - \hat{g}_a}] = \sum_{a} p_a$, and $\textbf{Var}[ \sum_{a} \abs{g_a - \hat{g}_a}] = \sum_a p_a(1 - p_a) \leq N \eta(1 - \eta)$, recalling that $\eta < 1/2$.  By Chebyshev's inequality
\begin{align}
    \P\left(\left|\sum_a|g_a-\hat{g}_a|- \sum_{a} p_a \right| > N \epsilon_1\right) \leq \frac{\eta(1-\eta)}{N \epsilon_1^2}, \label{eq:misclassification_bound}
\end{align}
so that we can bound $\frac{1}{N}\sum_a|g_a-\hat{g}_a|$ with high probability by $\epsilon_1 + \eta$.  Note that by the same logic above, the second and third terms of Eq.~\eqref{eq:erm_decomposition} can be bounded together by observing that their sum is bounded by $$\left(\frac{RN}{N_{S^*}} + \frac{RN}{N_{\hat{S}}}\right) \left(\frac{1}{N} \sum_{a} \abs{g_a - \hat{g}_a} \right)$$ 

\textbf{Bounding the fourth term}: Consider $G^*: \cA \rightarrow \{0, 1\}$ to be fixed, and let $\cS$ be our hypothesis class of functions $S: \cX \rightarrow \{0, 1\}$.  
For any $S: \cX \rightarrow \{0, 1\}$, define $f(Z) \coloneqq (Y - \E[Y \mid X]) \cdot G^*(A) \cdot S(X)$, then (defining $n$ as $NR$, our total number of samples)
\begin{align*}
  \hat{Q}(\hat{S}, G^*) &= \frac{1}{\hat{\P}(\hat{S})} \left(\frac{1}{n} \sum_{i=1}^{n} f(Z_i)\right), &\text{and}&& Q(\hat{S}, G^*) &= \frac{1}{\P(\hat{S})} \E[f(Z)],
\end{align*}
where to deal with the fact that $\P(S) \neq \hat{P}(S)$, we can write that 
\begin{align*}
  &\hat{Q}(\hat{S}, G^*) - Q(\hat{S}, G^*) \\
  &= \left(\frac{1}{\hat{\P}(\hat{S})} - \frac{1}{\P(\hat{S})}\right) \frac{1}{n} \sum_{i=1}^{n} f(Z_i) + \frac{1}{\P(\hat{S})} \left(\frac{1}{n} \sum_{i=1}^{n} f(Z_i) - \E[f(Z)]\right) \\
  &\leq \abs{\frac{1}{\hat{\P}(\hat{S})} - \frac{1}{\P(\hat{S})}} + \frac{1}{\P(\hat{S})} \left(\frac{1}{n} \sum_{i=1}^{n} f(Z_i) - \E[f(Z)]\right),
\end{align*}
and we can bound the last term by standard learning theory arguments.  In particular, this can be seen as a weighted loss, where $f(Z) = W(Z) \cdot S(Z)$, for $W \coloneqq (Y - \E[Y \mid X]) \cdot G(A)$.  Note that each $S \in \cS$ defines some $f \in \cF$. In particular, we can write that with probability at least $1 - \delta_1$, for all $f \in \cF$, we have
\begin{align}
  \sup_{f \in \cF} \left[ \frac{1}{n} \sum^{n}_{i=1} f(Z_i) - \E f(Z) \right] &\leq 2 \cR(\cF, n) + 4 \sqrt{(2 \ln (4/\delta_1))/n} \nonumber \\
  &\leq 2 \cR(\cS, n) + 4 \sqrt{(2 \ln (4 / \delta_1)) / n}  \label{eq:rademacher_bound}
\end{align}
where we have used the fact that $\abs{f(Z)} \leq 1$, and where $\cR$ is the Rademacher complexity \citep[Thm. 26.5.2 of][]{Shalev-Shwartz2014}.  In the second line, we use the fact that $\cF = \phi \circ S$, where $\phi(S) = W \cdot S$ is a 1-Lipschitz function, since $\abs{W} \leq 1$.  By the contraction lemma \citep[Lemma 26.9 of][]{Shalev-Shwartz2014}, we have it that $\cR(\cF, n) \leq \cR(\cS, n)$.

\textbf{Combining bounds}: By the union bound, the bounds in Eq.~\eqref{eq:concentration_bound}, Eq.~\eqref{eq:misclassification_bound} and Eq.~\eqref{eq:rademacher_bound} hold with probability at least $1-\delta$, where
\begin{equation}
\label{eq:high_probability}
    \delta=\exp\left(-\frac{\beta'NR\epsilon_0^2}{2}\right)+\frac{\eta(1-\eta)}{N\epsilon_1^2} + \delta_1.
\end{equation}
Hence, with probability at least $1 - \delta$, we can bound Eq.~\eqref{eq:erm_decomposition} by
\begin{align*}
&Q(S^*, G^*)-Q(\hat{S}, G^*) \le \epsilon_0 + \left(\frac{1}{\beta'}+\frac{1}{\beta}\right)(\eta+\epsilon_1) + \abs{\frac{1}{\hat{\P}(\hat{S})} - \frac{1}{\P(\hat{S})}} \\
&\quad \quad + (1 / \P(\hat{S}))\left( 2 \cR(\cS, NR) + 4 \sqrt{(2 \ln (4 / \delta_1)) / NR} \right),
\end{align*}
where we have used $NR / N_{S^*} = 1 / \beta'$ and $NR / N_{\hat{S}} = 1 / \beta$, by definition of $\beta'$ and because $\hat{S}$ is chosen to be a $\beta$-fraction of the dataset.  Finally, we simplify by setting the three terms in Eq. \ref{eq:high_probability} to be equal, and expressing $\epsilon_0$ and $\epsilon_1$ in terms of $\delta$:
\[
    \epsilon_0=\sqrt{\frac{2\ln(3/\delta)}{\beta'NR}}, \ \ \epsilon_1=\sqrt{\frac{3\eta(1-\eta)}{\delta N}}, \ \ \delta_1=\delta/3
\]
from which the desired result follows.
\end{proof}

\section{Semi-Synthetic Experiments: Additional Details}
\label{sec:semi-synth-details-app}

\subsection{Semi-Synthetic Setup Details}
\label{sec:semisynth-setup-app}
\textbf{Additional Dataset Details}: The Stanford study on human predictions of recidivism only included participants who passed both attention checks in the assessment. Participants were provided information about the baseline recidivism rates and charge. We use the responses from participants who were not given feedback so that decisions from the same participant are independent and identically distributed. The participants provide probabilistic predictions, which we convert into binary decisions by thresholding at 50\%. By construction of the survey, no participant can predict exactly 50\%.

\textbf{Semi-Synthetic Policy Generation (Details)}: For both alternative policies (as logistic regression models), we add a new binary feature corresponding to the binary region indicator (1 if the sample falls in the region, 0 otherwise).  The weight of this new coefficient is 1.5 for the alternative policy and 0 for the base policy.  We give additional details on the differences between the base and alternative policies below:
\begin{itemize}
    \item \textit{Drug Possession}: On average, the base and alternative policies predict recidivism 56\% and 61\% of the time, respectively, outside the subset of drug-possession charges. Within the subset, they predict recidivism 60\% and 87\% of the time, respectively.  
    \item \textit{Misdemeanor and Age at Most 35}: On average, the base and alternative policies predict recidivism 56\% and 62\% of the time, respectively, outside the subset of misdemeanor charges for individuals with age at most 35. Within the subset, the averages are 48\% and 79\% respectively.  
\end{itemize}

\subsection{Baseline Details}%
\label{sec:baseline_app}
In this section, we describe implementation details for the two baselines discussed in section \ref{sub:performance_versus_baselines}, two additional baselines we created, limitations of the baselines, and additional evaluation on the semi-synthetic set-ups.

\textbf{Direct Model:} The agents are included in a one-hot encoding, with the last agent dropped to prevent co-linearity. The agents are partitioned into two equal-size groups based on their logistic regression coefficients. Logistic regressions (LR), ridge regressions (RR), decision trees (DT), and random forests (RF) are via scikit-learn \citep{scikit-learn}. For logistic regressions, we tune the L2 regularization constant ($C$) among 10, 1, 0.1, 0.01, 0.001, 0.0001, and 0.00001. For ridge regressions, we tune the L2 regularization constant ($\alpha$) among 0.01, 0.1, 1, 10, and 100. For decision trees and random forests, we tune the minimum samples per leaf among 10, 25, and 100. For random forests, we also tune the number of trees among 10, 25, and 100.

\textbf{TARNet}: Our optimizer performs stochastic gradient descent with zero momentum on the binary cross entropy loss function from PyTorch \citep{paszke2019pytorch}. We tune the learning rate among 0.0001, 0.001, and 0.01 for the drug possession set-up and among 0.0001, 0.0005, 0.001, 0.005, and 0.01 for the misdemeanor set-up. We tune the number of units in each of the layers among 10 and 20 for the drug possession set-up and among 10, 20, and 30 for the misdemeanor set-up. Finally, we tune dropout among 0.05 and 0.25. We train the model for 200 epochs and use the epoch that has the lowest validation loss.

\textbf{U-learners:} A standard U-learner \citep{nie2017quasi} is designed to estimate the effect of a binary treatment by predicting treatment from features $T = f\left(X\right)$, predicting outcome from features $Y = g\left(X\right)$, and then predicting the ratio between the residuals of the previous two models from features $\left(Y - \hat{g}\left(X\right)\right)/\left(T - \hat{f}\left(X\right)\right) = h\left(X\right)$. To adapt this model to multiple discrete treatments, we fit $f_{ij}$ for every pair of agents $A_i$ and $A_j$ using only the samples from those two agents. This means there is a separate $h_{ij}$ for each pair of agents.  The amount of variation for a sample $x$ is defined as $\sum_{i,j\in\cA} \vert\hat{h}_{ij}(x)\vert$. As in the direct models, a model is fit to predict this quantity, and the top $\beta$ quantile of these predictions is used to identify region membership. To create $\hat{G}$, we first identify a pair of agents that are most dissimilar using the sum of absolute values of the difference between that pair across all samples within the region. All the other agents are added to the partition based on which provider in the starting pair they are closer to.

\textbf{Causal Forest Adaptations:} A causal forest \citep{wager2018estimation} resembles a random forest with the splits chosen instead to maximize the difference in the effect of the treatment on outcome. A causal forest can only predict the difference between two treatments. The na\"ive adaptation of fitting $O\left(\vert \cA \vert^2\right)$ causal forests between each pair of agents becomes computationally infeasible when $\vert \cA \vert$ is large. Instead, we use the following heuristic to learn a partition with causal forests: Initialize the groupings $G_0$ and $G_1$ with a few agents based on the U-learner predictions. We start with the most dissimilar pair as defined above. Then, if a provider is much closer to one of the two starting providers (defined as the sum of absolute differences being at most a tenth of the sum for the other pair), add that provider to the group with the closer provider. To add agent $a$, compute a causal forest for the difference in treatment choice between $G_0 \cup a$ and $G_1$ and the difference between $G_0$ and $G_1 \cup a$. We use the EconML implementation of causal forests \citep{econml}. For hyperparameter tuning, predictions $U_{a_0, a_1}\left(X\right)$ from the U-learner are used as "oracle" predictions. The "oracle" predictions between groups $G_0, G_1$ are defined as $\sum_{a_0 \in G_0, a_1 \in G_1}  U_{a_0, a_1}\left(X\right)$. We compare each pair of causal forests and add $a$ to $G_0$ if the former has a larger sum of the absolute difference across all samples. Otherwise, add $a$ to $G_1$. Repeat until all agents have been added to a group. A region model is applied to the predictions from the final causal forest. 

\textbf{Hyperparameter Tuning:} For both the baselines and our algorithm, we tune the hyperparameters for each piece of the model separately, e.g. each decision tree piece of the model is tuned separately. The data is split into train, validation, and test 60/20/20 stratified by provider, guaranteeing at least 1 sample per provider in each of the sets. The same validation set is used for all steps in a single model. After the best hyperparameter is selected, the model is retrained on both the training and validation data.

\textbf{Baseline Limitations:} A limitation of the TARNet baseline is that it does not learn a partition of providers. A limitation of the other baselines except for the U-learner is that the provider partition is based on all samples, not just samples within the region.

\begin{figure}[t]
\centering
  \includegraphics[width=\textwidth]{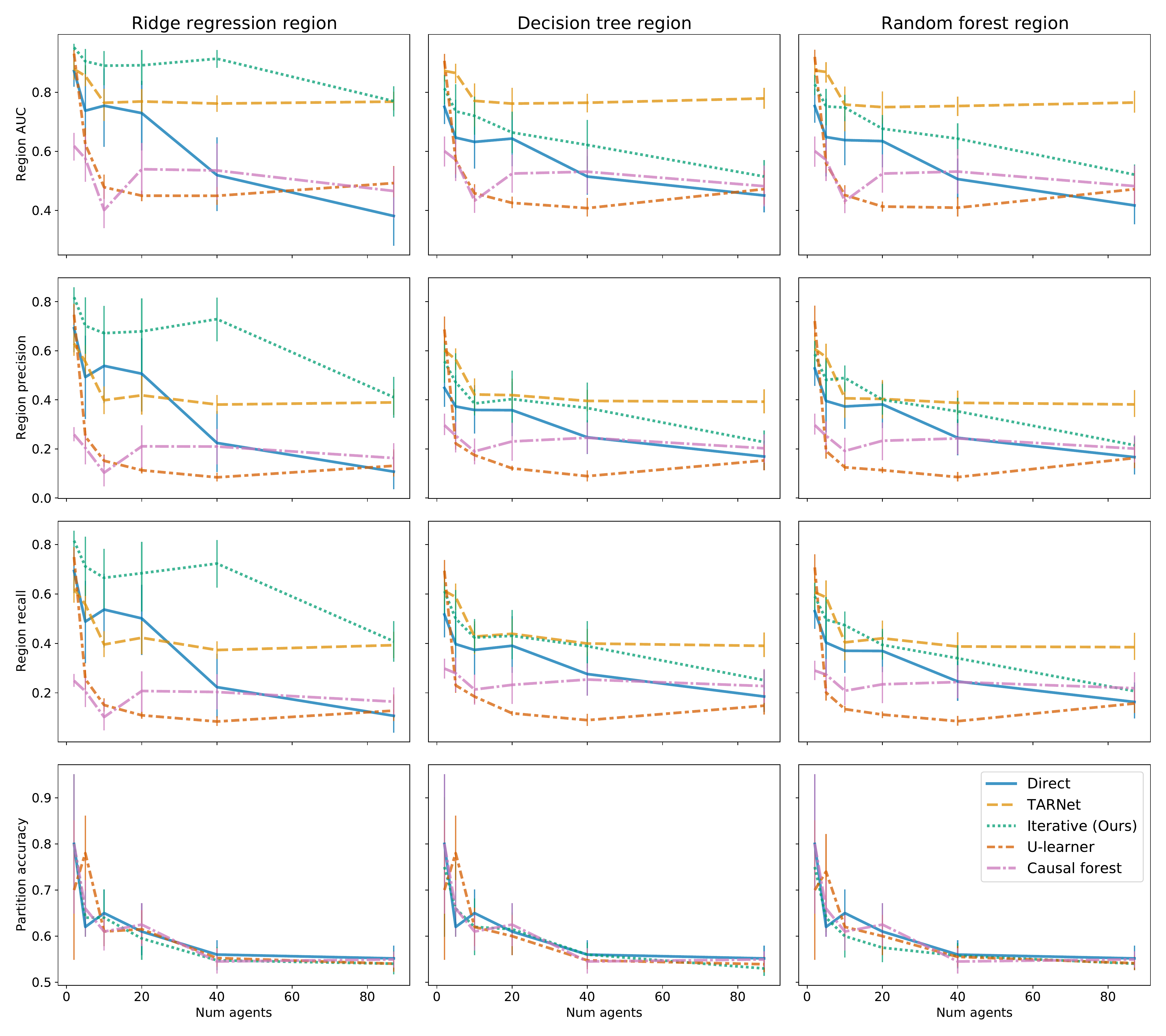}
  \caption{Comparison to baseline approaches for drug possession semi-synthetic set-up. Uncertainty bands represent 95\% intervals for the mean derived via bootstrapping computed using seaborn \citep{Waskom2021seaborn}. Our method with a ridge regression region model is the best at identifying the region across all combinations of outcome and region models. 
  }%
  \label{fig:semisynth_drug_app}
\end{figure}

\begin{figure}[t]
\centering
  \includegraphics[width=\textwidth]{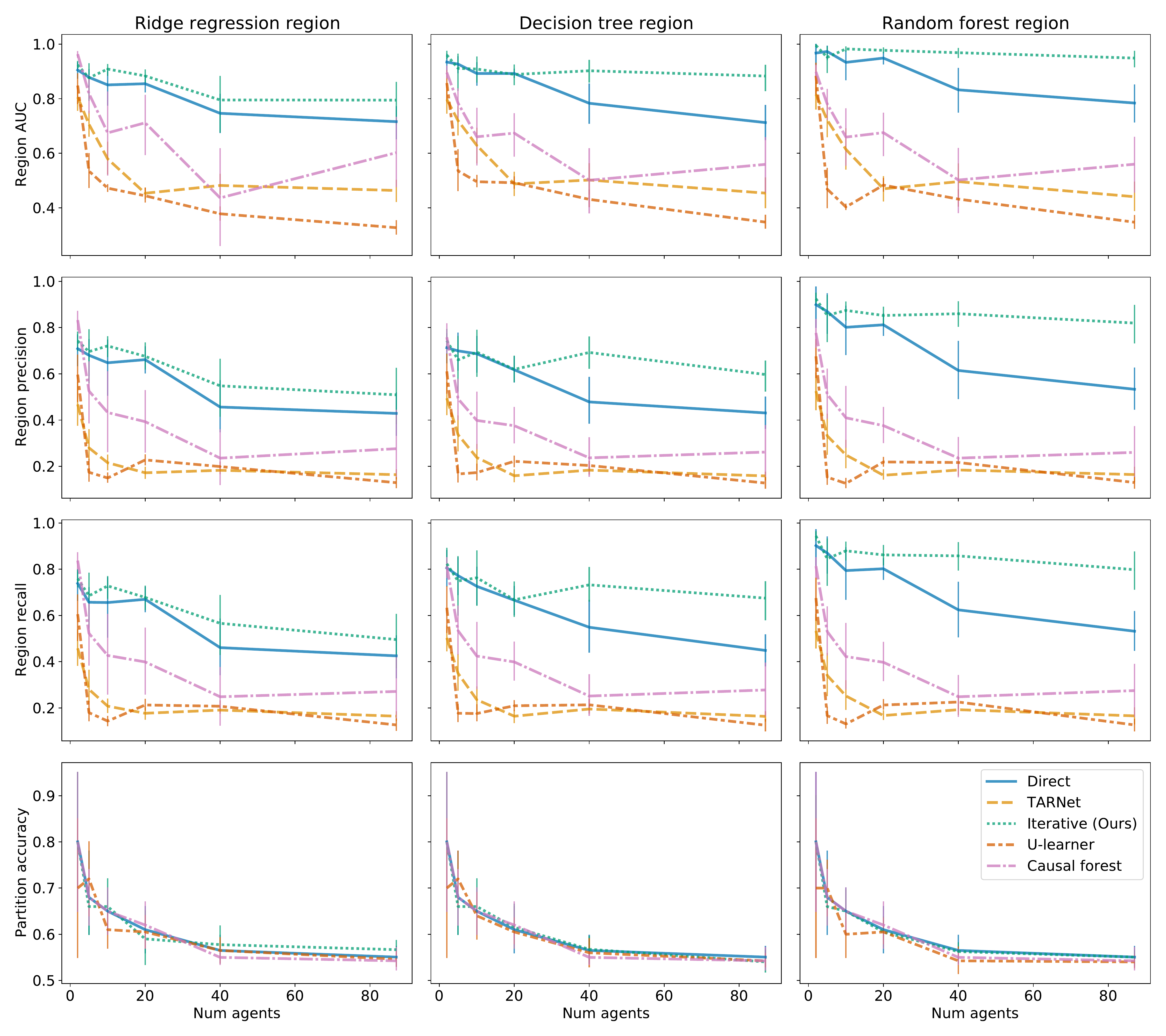}
  \caption{Comparison to baseline approaches for misdemeanor semi-synthetic set-up. Uncertainty bands represent 95\% intervals for the mean derived via bootstrapping computed using seaborn \citep{Waskom2021seaborn}. Our method with a random forest region model is the best at identifying the region across all combinations of outcome and region models. 
  }%
  \label{fig:semisynth_misdemeanor_app}
\end{figure}

\textbf{Additional Baseline Results:} Other metrics we consider are the precision and recall of the region and the accuracy of the partition. To compute accuracy, we compare the learned groups 0 and 1 with the true groups 0 and 1, respectively, and with the true groups 1 and 0, respectively, and take the higher of the two. Figure \ref{fig:semisynth_drug_app} shows that the ridge regression model we highlight in the main paper fits the region better than the other two models for large numbers of agents for the drug possession set-up. On the other hand, for the misdemeanor under age 35 set-up, Figure \ref{fig:semisynth_misdemeanor_app} shows that the random forest model fits the region better. TARNet is the best-performing baseline for the drug possession set-up, while the direct model is the best-performing baseline for the misdemeanor set-up. Our model outperforms both in all cases. The partition is difficult for any model to learn in both set-ups when the number of agents increases. Despite this challenge, our algorithm is still able to recover the region reasonably well.

\subsection{Convergence Analysis}
\label{conv-semisynth-app}
\begin{figure}[t]
\centering
  \includegraphics[width=\textwidth]{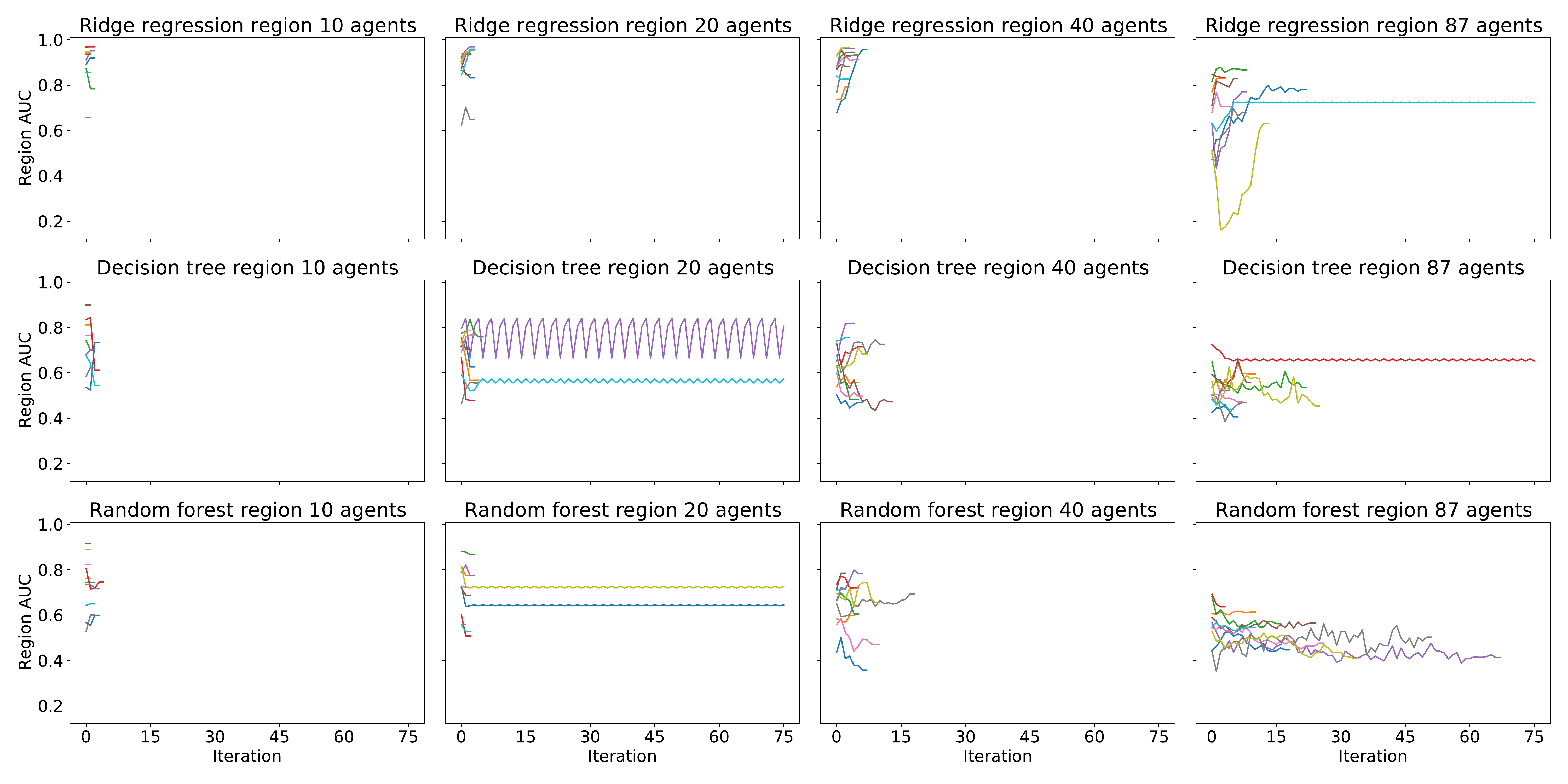}
  \caption{Convergence of iterative algorithm in drug possession semi-synthetic set-up. Each of the 10 lines in each plot represents a dataset generated from a different random seed. Region AUC is computed after the partition is updated. Algorithm terminates when partition does not change. Although the iterative algorithm was run for up to 100 iterations, the number of iterations plotted was truncated to show all terminations. 2 and 5 agents are omitted because they closely resemble the plots for 10 agents.
  }%
  \label{fig:semisynth_drug_conv_app}
\end{figure}

\begin{figure}[t]
\centering
  \includegraphics[width=\textwidth]{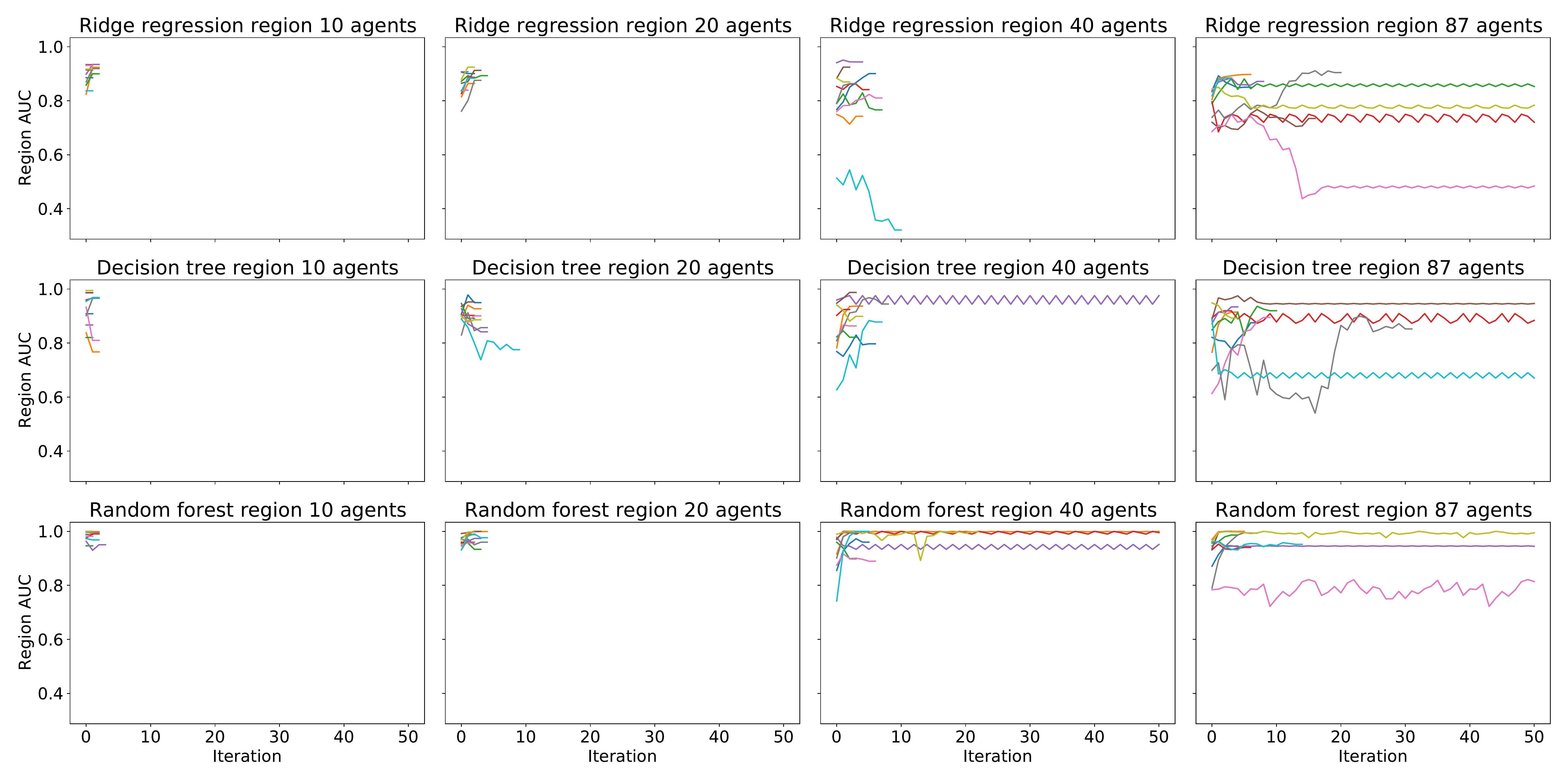}
  \caption{Convergence of iterative algorithm in misdemeanor semi-synthetic set-up. See Figure \ref{fig:semisynth_drug_conv_app} for description.
  }%
  \label{fig:semisynth_misdemeanor_conv_app}
\end{figure}

We derive a generalization bound on the performance of the iterative algorithm after one iteration in Section \ref{sec:algorithm_analysis}. Here we examine empirical performance across additional iterations. To do so, we compute the AUC of the region in each iteration and how many iterations the algorithm needs to converge. Note that because AUC is not the objective that we are optimizing, there are no guarantees that it increases monotonically across iterations. As seen in Figures \ref{fig:semisynth_drug_conv_app}-\ref{fig:semisynth_misdemeanor_conv_app}, in both semi-synthetic set-ups, for small numbers of agents, convergence is almost immediate. For larger numbers of agents, the algorithm almost always converges. The exceptions are where the algorithm enters what appears to be a cycle. This problem can likely be addressed by re-initializing the model if the algorithm is stuck in a cycle.

\subsection{Robustness Analysis}
\label{app:robustness-semi}
\begin{figure}[t]
\centering
  \includegraphics[width=\textwidth]{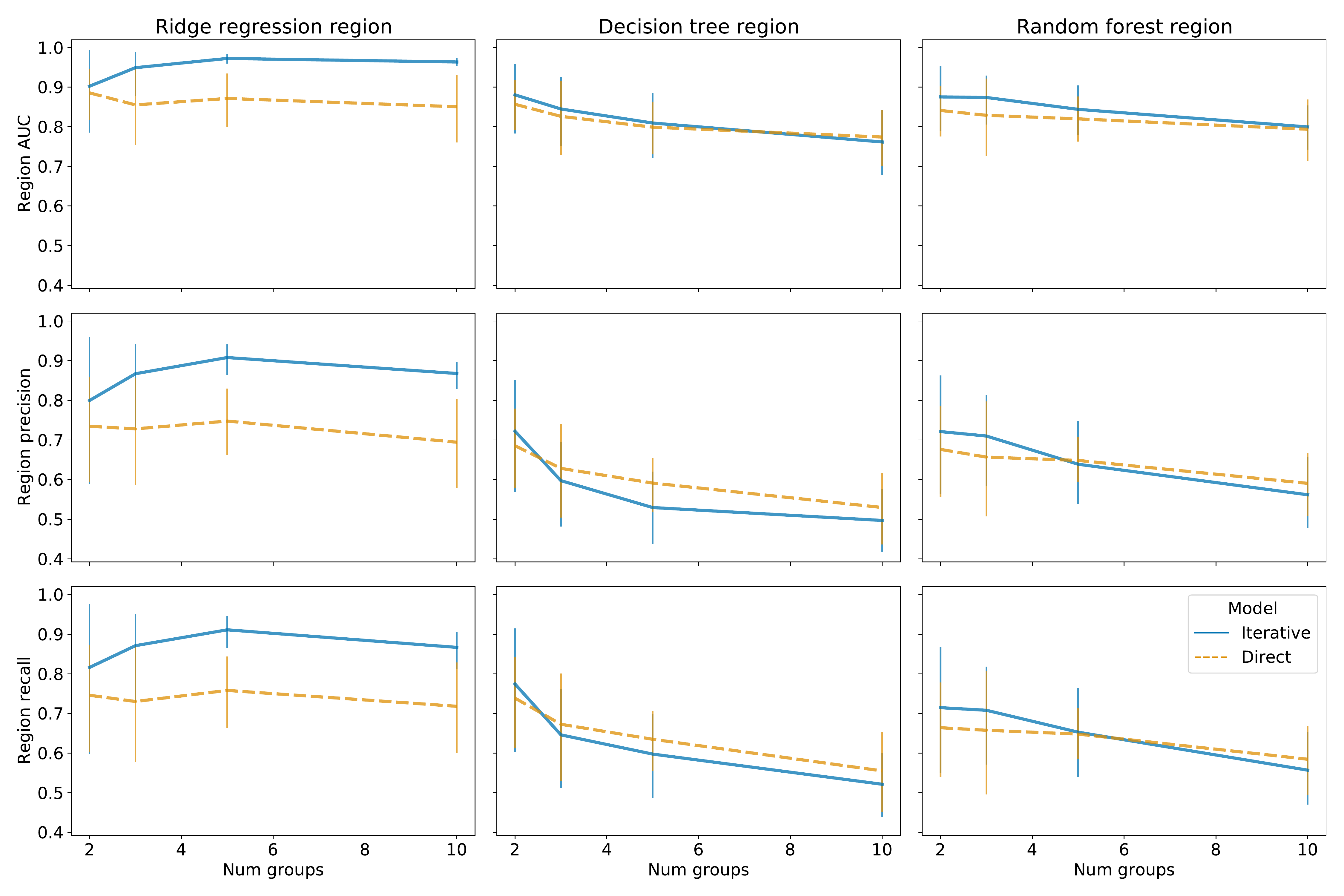}
  \caption{Robustness to the assumption that there are 2 agent groups in the drug possession semi-synthetic set-up. The 40 agents in each set-up are roughly equally divided into 2, 3, 5, and 10 groups. Uncertainty bands represent 95\% intervals for the mean derived via bootstrapping computed using seaborn \citep{Waskom2021seaborn}.
  }%
  \label{fig:semisynth_drug_robustness_app}
\end{figure}

\begin{figure}[t]
\centering
  \includegraphics[width=\textwidth]{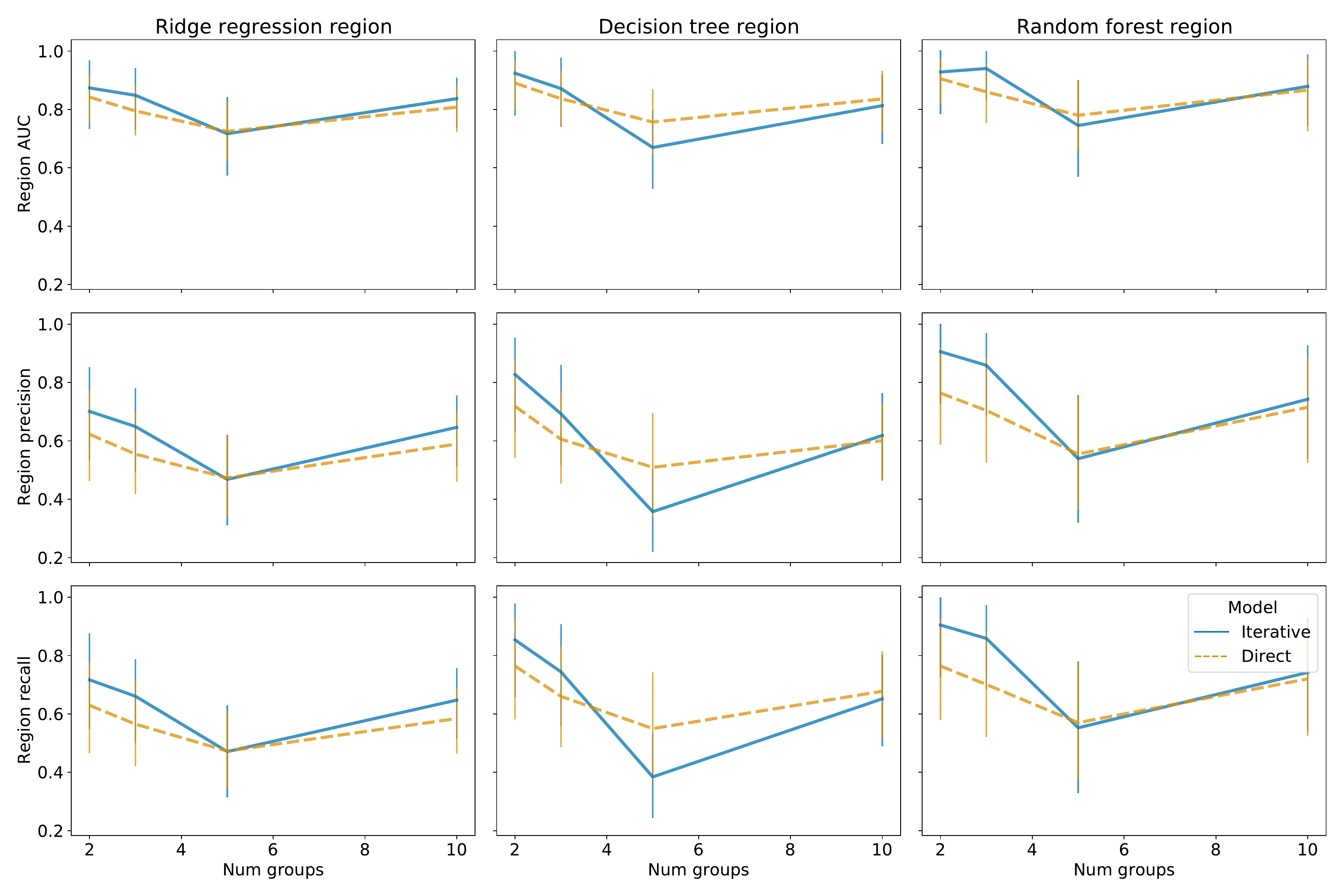}
  \caption{Robustness to the assumption that there are 2 agent groups in the misdemeanor under age 35 semi-synthetic set-up. See Figure \ref{fig:semisynth_drug_robustness_app} for description.
  }%
  \label{fig:semisynth_misdemeanor_robustness_app}
\end{figure}

We examine robustness of the model to violations of the assumption that there are 2 agent groups. For instance, it may be more realistic for agents to have a wide spectrum of preferences. We represent this in our semi-synthetic set-up by varying the coefficient on the region indicator variable described in Appendix \ref{sec:semisynth-setup-app} in equally spaced steps between -1.5 and 1.5. The number of agents is held constant at 40. The region and the policy outside the region are the same as before. This corresponds to varying the average predicted probabilities of recidivism within the subset in the groups between around 30\% and 86\% for the drug possession set-up and between around 17\% and 79\% for the misdemeanor under age 35 set-up. We compare our method with the direct baseline with all region models. As seen in Figure \ref{fig:semisynth_drug_robustness_app}, the ridge regression model is most robust to this violation in the drug possession set-up. In fact, performance actually improves when the number of agent groups increases. We hypothesize this may occur for two reasons: 1. The ridge regression model parametrizes the region best, as that was also the globally best model in the 2-group specification. 2. The agents can still be divided into two groups based on above and below average preference, so this violation of the 2-group assumption is relatively simple. Figure \ref{fig:semisynth_misdemeanor_robustness_app} shows that all 3 region models are somewhat robust to increasing numbers of groups for both the iterative and direct baseline in the misdemeanor under 35 set-up.

\clearpage
\section{Diabetes Experiment: Additional Details}%
\label{sec:diabetes_app}

\textbf{Data Details, Train/Validation/Test Split}: The data set of de-identified health insurance claims was provided by a large insurance company. That company obtained the relevant consent from individuals to use their data, and gave us permission to use the data for research purposes. A waiver of informed consent was obtained in compliance of HIPAA.

We require at least 3 years of observation before the first diabetes treatment to ensure that the observed treatment is indeed first-line. We also require at least one diagnosis code of diabetes mellitus and at least one A1C measurement at least 6.5 prior to first-line treatment. Patients who had at least one diagnosis code related to type 1 diabetes mellitus prior to first-line treatment or gestational diabetes (pregnancy, neonatal diabetes, or diabetes of the young) in the 1 year prior to first-line treatment are excluded. The exclusion of gestational diabetes is important to prevent confounding since those patients may see specialized providers and receive different treatments. Patients who received more than one first-line treatment or any treatment besides metformin, DPP-4 inhibitors, or sulfonylureas are also excluded.

We only include agents with at least 4 samples. For each agent, 37.5\% of samples are placed in the training set, 12.5\% in validation, and 50\% in test to ensure the test set is sufficiently large for computing the partition in $L\left(\hat{S}\right)$ on the test set. We require at least one sample per agent in each of the training and validation sets and two in the test set. Treatment date is converted to seconds, and all features are normalized to have mean 0 and standard deviation 1.

\begin{figure}[t]
\centering
  \includegraphics[width=.6\textwidth]{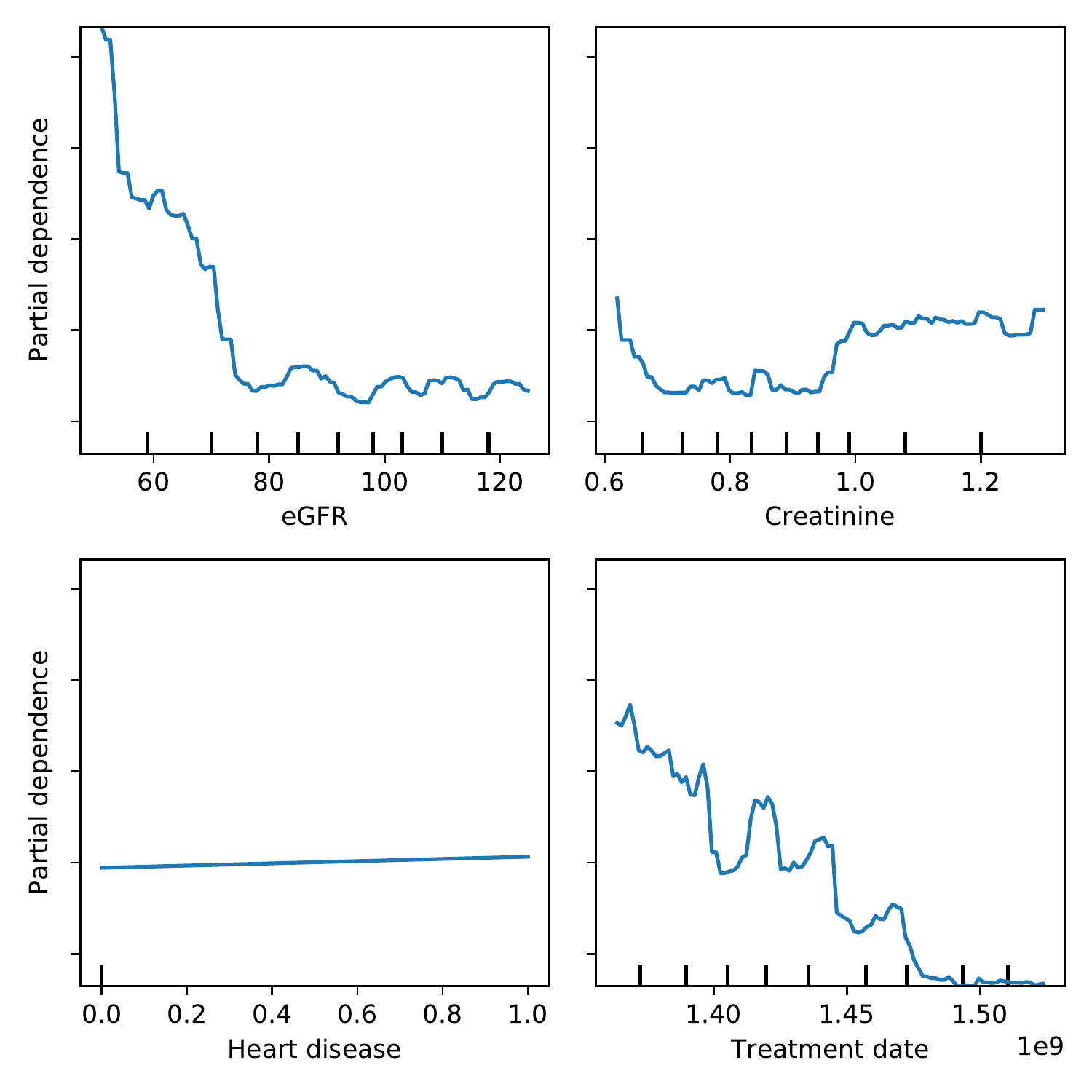}
  \caption{Partial dependence plot of random forest outcome model in diabetes experiment
  }%
  \label{fig:diabetes_pdp}
\end{figure}

\textbf{Selecting the outcome model}: For the outcome model, we consider three hypothesis classes: logistic regressions, decision trees, and random forests. The models are tuned as described in Appendix~\ref{sec:baseline_app}. We select a random forest for three reasons: \begin{enumerate*} \item Random forests have the highest AUC on the validation set, as shown in Table \ref{tab:diabetes_outcome_model_aucs}. \item Random forests are better calibrated in the discovered region, as shown in Table \ref{tab:diabetes_outcome_model_calibration}. \item The partial dependence plots in Figure \ref{fig:diabetes_pdp} show reasonable relations between each feature and the outcome.\end{enumerate*} These results are reported for the fold that was selected based on significance and calibration statistics.

\begin{table}[h]
    \centering
    \caption{Validation AUCs for outcome model for diabetes experiment.}
        \begin{tabular}{lc}
    \toprule
        \textbf{Model} & \textbf{Test AUC} \\
        \midrule
        Logistic regression & 0.6957 \\
        Decision tree & 0.7351 \\
        Random forest & 0.8283 \\
        \bottomrule
        \end{tabular}
    \label{tab:diabetes_outcome_model_aucs}
\end{table}

\begin{table}[h]
    \centering
    \caption{Calibration of outcome model on regions selected with each outcome model for diabetes experiment. Comparison of average true and predicted outcomes in region among training samples.}
        \begin{tabular}{lcc}
    \toprule
        \textbf{Model} & \textbf{True average} & \textbf{Predicted average} \\
        \midrule
        Logistic regression & 0.2485 & 0.1839 \\
        Decision tree & 0.2023 & 0.1688 \\
        Random forest & 0.2343 & 0.2226 \\
        \bottomrule
        \end{tabular}
    \label{tab:diabetes_outcome_model_calibration}
\end{table}

\begin{figure}[t]
  \begin{subfigure}[h]{0.6\textwidth}
  \centering
    \tiny
    \begin{tikzpicture}[sibling distance=10em, level distance=7em,
        main node/.style = {shape=rectangle, rounded corners,
            draw, align=center,
            top color=white, bottom color=blue!20},
        edge node/.style = {shape=rectangle, rounded corners, align=center,
            top color=white, bottom color=white},
        var node/.style = {shape=rectangle, rounded corners,
            draw, align=center,
            top color=white, bottom color=red!50},
        lessvar node/.style = {shape=rectangle, rounded corners,
            draw, align=center,
            top color=white, bottom color=red!50},
        level 1/.style={sibling distance=11em},
        level 2/.style={sibling distance=9em},
        level 3/.style={sibling distance=11em},
        level 4/.style={sibling distance=12em}]
        \node(0)[main node]{\textbf{Node 0}\\ eGFR $>$ 71.5?}
            child { node [var node]{\textbf{Node 1}\\ $Q(S, G) = 0.097$ \\ G=0 MET: 169/205 (82\%) \\ G=1 MET: 110/213 (52\%)}
                edge from parent node[edge node]{No}
            }
            child { node [main node]{\textbf{Node 2}\\ Creatinine > 0.815?}
                child { node [main node]{\dots}
                    edge from parent node[edge node]{No}
                }
                child { node [main node]{\textbf{Node 6}\\ eGFR $>$ 98.5?}
                    child { node [main node]{\textbf{Node 7}\\ Creatinine $>$ 0.995?}
                        child { node [main node]{\textbf{Node 8}\\ $Q(S, G) = 0.002$\\ G=0 MET: 114/126 (90\%)\\ G=1 MET: 97/105 (92\%)}
                            edge from parent node[edge node]{No}
                        }
                        child { node [main node]{\textbf{Node 9}\\ $Q(S, G) = 0.000$\\ G=0 MET: 83/93 (89\%)\\ G=1 MET: 82/96 (85\%)}
                            edge from parent node[edge node]{Yes}
                        }
                        edge from parent node[edge node]{No}
                    }
                    child { node [lessvar node]{\textbf{Node 10}\\ $Q(S, G) = 0.049$\\ G=0 MET: 113/117 (97\%)\\ G=1 MET: 69/88 (78\%)}
                        edge from parent node[edge node]{Yes}
                    }
                    edge from parent node[edge node]{Yes}
                }
                edge from parent node[edge node]{Yes}
            }
        ;
    \end{tikzpicture}
    \caption{}
    \label{fig:diabetes_IterativeAlgDecisionTree_DecisionTree_fold2}
  \end{subfigure}%
    \begin{subfigure}[h]{0.4\textwidth}
    \centering
    \includegraphics[width=\linewidth]{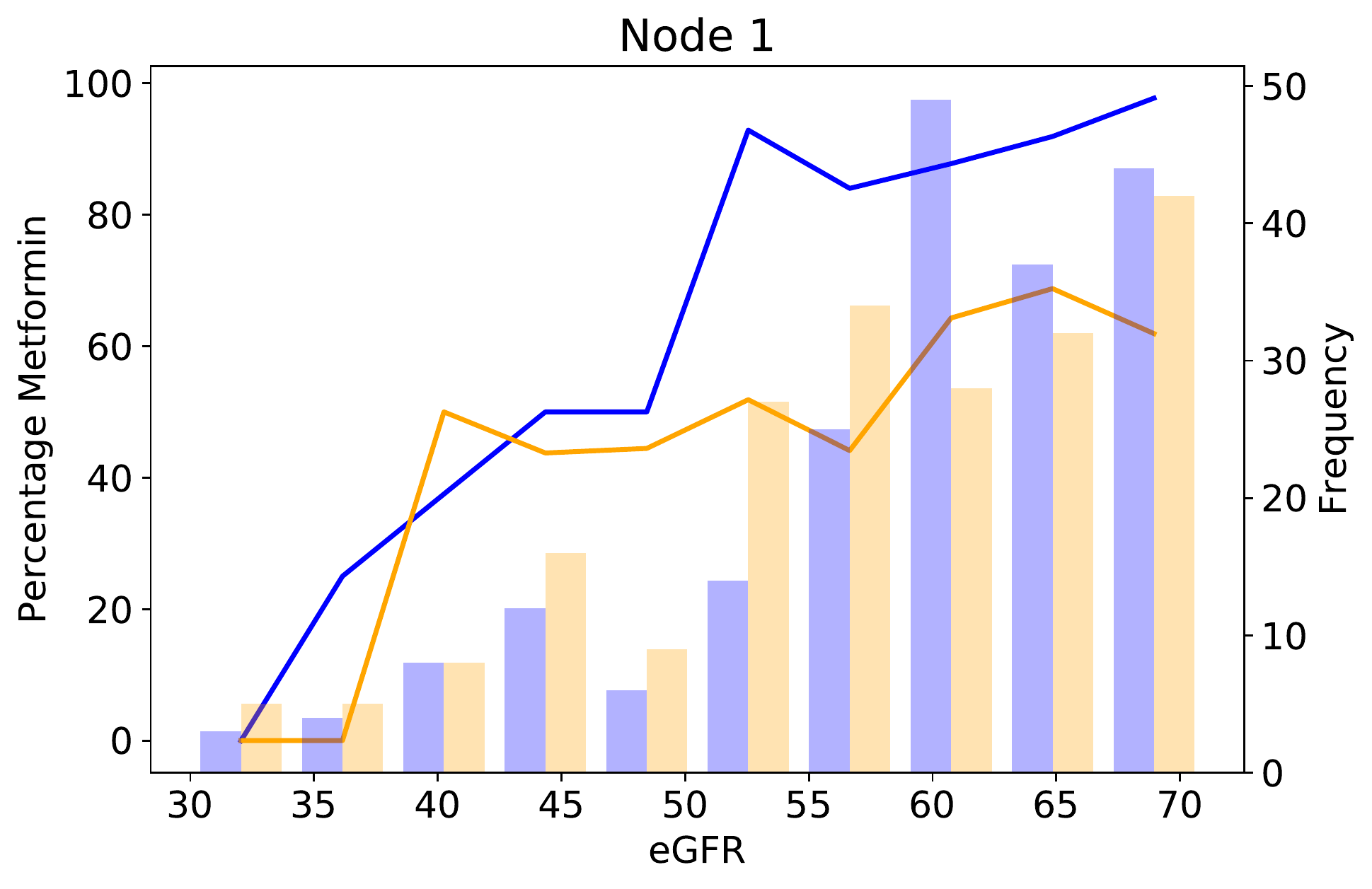}
    \includegraphics[width=\linewidth]{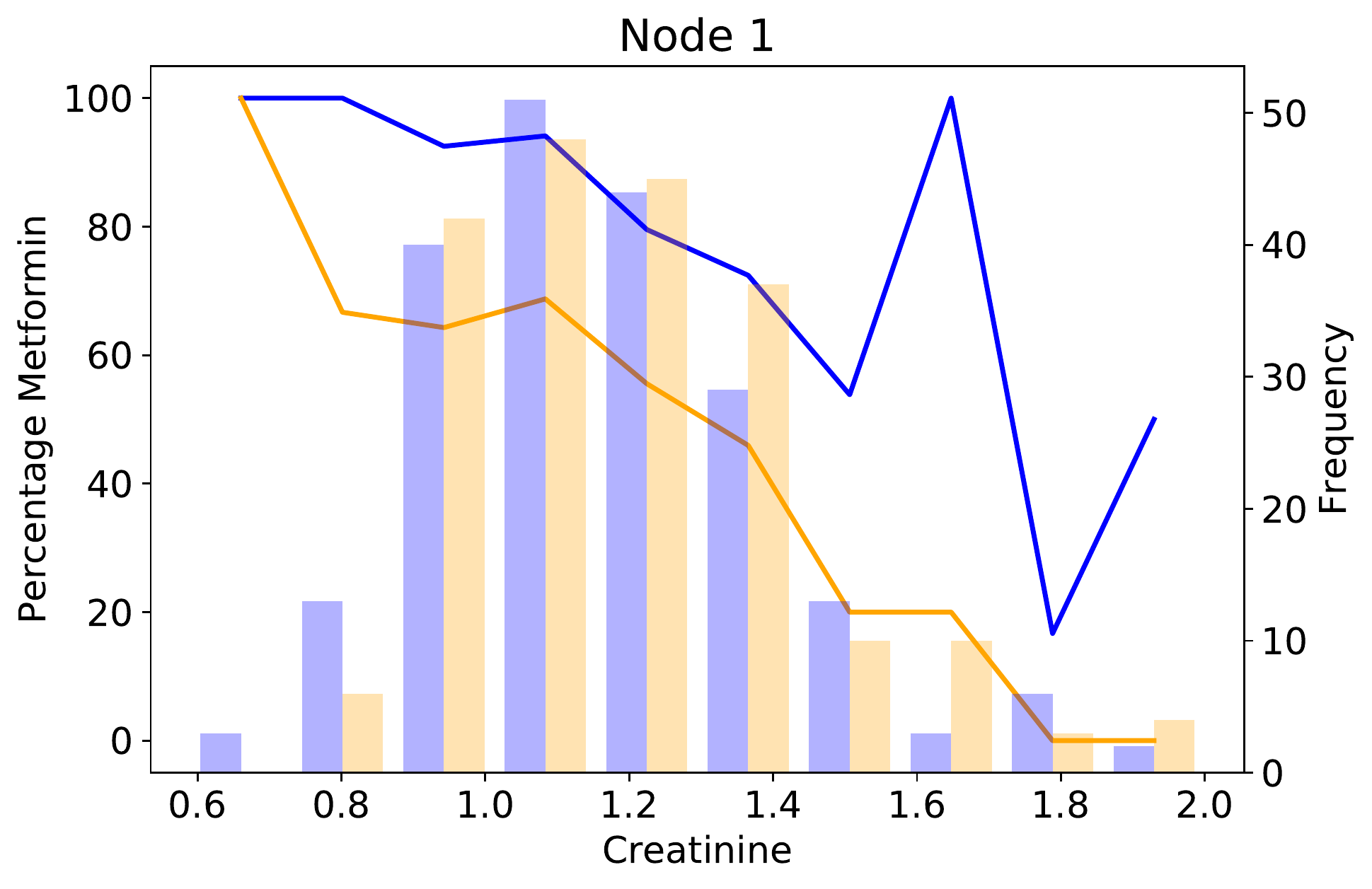}
    \caption{}
    \label{fig:diabetes_IterativeAlgDecisionTree_region_variation}
\end{subfigure}
\caption{(\subref{fig:diabetes_IterativeAlgDecisionTree_DecisionTree_fold2}) Decision tree identifying regions of disagreement for first-line treatment decisions in diabetes, indicated in red. MET\@: Metformin. G=0 denotes the group found to prefer metformin as an initial treatment, and G=1 indicates the group with the opposing preference. Numbers at leaves computed on a held-out test set. $Q(S, G)$ values are the $L\left(\hat{S}\right)$ metric computed on the test set; number of samples are computed using only providers with at least two samples in the region of heterogeneity. (\subref{fig:diabetes_IterativeAlgDecisionTree_region_variation}) Variation in Node 1 of the tree in Figure \ref{fig:diabetes_IterativeAlgDecisionTree_DecisionTree_fold2}, one of the regions of disagreement in the diabetes dataset. The colors denote group membership: blue is $G = 0$, and orange is $G = 1$. The lines indicate the proportion of decisions in each bin where the agent prescribed metformin. The gap between the lines illustrates that the group-specific bias towards metformin generally holds across patient features. Only agents with at least two samples in the region are included. The histograms show the total number of samples in each bin. }
\label{fig:diabetes}
\end{figure}

\textbf{Visualizing the Region}: To describe the region in an interpretable way, we use a decision tree for the region model $h(x)$ described in Algorithm~\ref{alg:main} with $\beta = 0.25$ and a maximum of 5 iterations. The minimum number of samples per leaf for the decision tree region model is tuned among larger choices (50, 100, 200) for a more interpretable region. Our algorithm outputs a decision tree $h(x)$, shown in Figure \ref{fig:diabetes_IterativeAlgDecisionTree_DecisionTree_fold2} and a threshold of $b=0.0741$ in the training $Q(S,G)$ values. The region $S=\{x\in \cX; h(x)\geq b\}$ consists of the two nodes indicated in red. To verify the generalizability of this result, we use a held-out test set to compute the metric $L\left(\hat{S}\right)$. In the test set, this metric is also greater on these two nodes than on all other nodes in the decision tree.  

We zoom into the first region in Figure \ref{fig:diabetes_IterativeAlgDecisionTree_region_variation}. We only include decisions made by providers with at least two samples in that node. Providers in group $G=0$ prefer to initiate treatment with metformin, while providers in group $G=1$ prefer to initiate with sitagliptin, glipizide, glimepiride, or glyburide. This group-specific bias (blue vs. orange lines) generally holds across the range of GFR and creatinine values, indicating that this preference is not explained by the patient's features.

\textbf{Assessing Stability}: We also measure the stability of the region and grouping identified by our method by performing cross-validation. We split the training and validation data into 4 equally sized portions and assign 1 portion as the validation set. The only overlap between the 4 validation folds are samples that belong to agents with fewer than 4 observations in the training and validation set to ensure that all validation folds contain at least 1 sample per agent. (1) A region is stable if points that are selected for the region from most training folds are also selected when they are in the validation fold. If we look at the points that are in 1 validation fold, among the 461 points that are in the region for 2 to 3 training folds, 309 are also selected when they belong to the validation fold. (2) We also examine the consistency of the test region. With an average test region size of 553.75 points, 350.75 points are in the test region for at least 3 folds (each point in only 3 folds contributes weight 0.75). (3) We assess stability of the grouping by examining whether pairs of providers are consistently on the same or opposite sides of the grouping. Among the 12,181 pairs of providers that have at least 1 training, validation, or test sample in the region in at least 3 of the folds, 10,251 pairs have the same relationship in at least 3 of the folds. All 3 of these statistics suggest our algorithm arrives at similar regions and groupings regardless of how the training and validation samples are split.

\section{Additional Real-Data Experiment: First-Line Parkinson's Treatment}
\label{sec:extra_real_data_experiments_app}
\label{sec:parkinsons}

\textbf{Context and Data}: The Parkinson's Progression Markers Initiative (PPMI) is an observational study that follows Parkinson's patients starting within 2 years of diagnosis in their \textit{de novo} cohort \citep{marek2011parkinson}. The study collected data across the US, Europe, Israel, and Australia between 2010 and 2018. We examine decisions between the two most common first-line treatments, levodopa and rasagiline. Clinical trials are interested in assessing the effects of these treatments \citep{pd2014long}. For context, we include age, disease duration, and a motor assessment (the total from part II and III of the Movement Disorder Society Unified Parkinson's Disease Rating Scale (MDS-UPDRS)) \citep{goetz2007movement}. The features are normalized to have mean 0 and standard deviation 1. The cohort consists of 260 patients at 23 study sites, which we use as \enquote{agents} $A$.  Note that while treatment decisions are not made at study sites, these sites capture rough geographic locations across which there may be heterogeneity in treatment. We fix the outcome model to be a decision tree and the region size $\beta = 0.25$.

\begin{figure}[t]
  \centering
  \tiny
  \begin{tikzpicture}[sibling distance=12em, level distance=7em,
        main node/.style = {shape=rectangle, rounded corners,
            draw, align=center,
            top color=white, bottom color=blue!20},
        edge node/.style = {shape=rectangle, rounded corners, align=center,
            top color=white, bottom color=white},
        var node/.style = {shape=rectangle, rounded corners,
            draw, align=center,
            top color=white, bottom color=red!50},
        lessvar node/.style = {shape=rectangle, rounded corners,
            draw, align=center,
            top color=white, bottom color=red!50}]
        \node(0)[main node]{\textbf{Node 0}\\ Age $>$ 70.502?}
            child { node [main node]{\textbf{Node 1}\\ Disease duration \\$>$ 1.418 years}
                child { node [main node] {\dots}
                    edge from parent node[edge node]{No}
                }
                child { node [main node] {\textbf{Node 3}\\ Age $>$ 62.461}
                child {node [main node] {\textbf{Node 4}\\ 
                $Q(S, G) = 0.000$ \\
                G = 0 LEV: 1/6 (17\%)\\
                G = 1 LEV: 2/5 (40\%)
                }
                edge from parent node[edge node]{No}}
                child { node [lessvar node] {\textbf{Node 5}\\
                $Q(S,G) = 0.047$ \\
                G = 0 LEV: 2/6 (33\%)\\
                G = 1 LEV: 8/13 (62\%)}
                edge from parent node[edge node]{Yes}}
                edge from parent node[edge node]{Yes}}
                edge from parent node[edge node]{No}
            }
            child { node [var node]{\textbf{Node 2}\\ $Q(S, G) = 0.272$ \\
            G = 0 LEV: 0/7 (0\%)\\
            G = 1 LEV: 24/27 (88\%)}
                edge from parent node[edge node]{Yes}
            };
    \end{tikzpicture}
    \caption{Decision tree identifying regions of disagreement for first-line treatment decisions in Parkinson's, indicated in red. LEV: Levodopa. G=1 denotes the group found to prefer levodopa as an initial treatment, and G=0 indicates the group with the opposing preference. See Figure \ref{fig:diabetes_IterativeAlgDecisionTree_DecisionTree_fold2} for explanation.}
    \label{fig:ppmi_IterativeAlgDecisionTree_DecisionTree_fold1}
\end{figure}
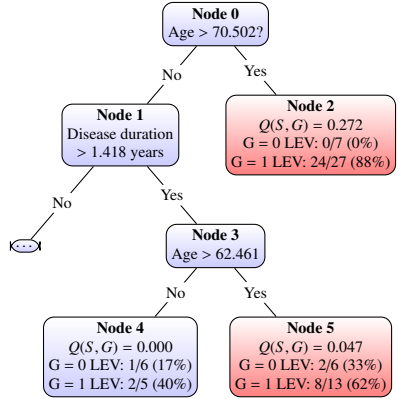

\textbf{Interpretation of Results}: The decision tree in Figure \ref{fig:ppmi_IterativeAlgDecisionTree_DecisionTree_fold1} shows that the selected region includes patients who are either above age 70 or both above age 62 and diagnosed at least around 1.5 years prior to first-line treatment. The region in the test set includes 13 patients in group 0 and 40 patients in group 1. Patients in the region in the test set are older on average (72.3 versus 64.3 for entire test set), have had Parkinson's longer (1.9 years versus 1.5 years) and have higher MDS-UPDRS part II + III scores (41.7 vs 37.5). Empirically, when stratified on age or MDS-UPDRS, patients in the top 33rd percentile (67 for age, 29 for MDS-UPDRS) are more likely to be given levodopa, while patients in the lower 67th are more likely to receive rasagiline. 

This region likely captures where treatment switches from rasagiline to levodopa. Our results align with clinical guidelines that levodopa may be better for patients with more motor and cognitive impairment due to fewer side effects \citep{muzerengi2015initial}, as these patients also tend to be older and have had the disease longer.

\textbf{Assessing Significance}: In Table~\ref{tab:IterativeAlgDecisionTree_benchmark_parkinsons}, we assess whether the region $S$ identifies variation in held-out data better than a randomly selected region. As in the diabetes experiment, we observe that the test statistics is close to the training statistic and more than 2 standard deviations from the average test statistic for random regions of the same size, suggesting the discovered region of heterogeneity generalizes beyond the training set.

\begin{table}[h]
    \centering
    \caption{Objective values $L(S)$ for the learned region on the training and test datasets, along with the distribution of values for randomly generated regions $S_{\text{rand}}$ given as mean (standard deviation). }
        \begin{tabular}{lll}
    \toprule
         \textbf{Metric} & \textbf{Subset} & \textbf{Value} \\
        \midrule
        $L(\hat{S})$ & Train & 0.2743 \\
        $L(\hat{S})$ & Test & 0.2170 \\
        $L(S_{\text{rand}})$ & Test & 0.1200 (0.0222) \\
        \bottomrule
        \end{tabular}
    \label{tab:IterativeAlgDecisionTree_benchmark_parkinsons}
\end{table}

\textbf{PPMI Disclaimer}: Data used in the preparation of this article were obtained from the Parkinson’s Progression Markers Initiative (PPMI) database (www.ppmi-info.org/data). For up-to-date information on the study, visit www.ppmi-info.org. PPMI – a public-private partnership – is funded by the Michael J. Fox Foundation for Parkinson’s Research and funding partners, including abbvie, AcureX therapeutics, Allergan, Aligning Science Across Parkinson's, Avid Radiopharmaceuticals, Bial Biotech, Biogen, BioLegend, Bristol Myers Squibb, Calico, Celgene, Dacapo brainscience, Denali, Edmond J. Safra Philanthropic Foundation, 4D Pharma PLC, GE Healthcare, Genentech, GlaxoSmithKline, Golub Capital, Handl Therapeutics, insitro, Janssen Neuroscience, Lilly, Lundbeck, Merck, Meso Scale Discovery, Neurocrine biosciences, Pfizer, Piramal, Prevail Therapeutics, Roche, Sanofi Genzyme, Servier, Takeda, Teva, ucb, verily, Voyager Therapeutics, and Yumanity Therapeutics.

\end{document}